\documentclass[lettersize,journal]{IEEEtran}
\usepackage{amsmath,amsfonts}
\usepackage{algorithmic}
\usepackage{algorithm}
\usepackage{array}
\usepackage[caption=false,font=normalsize,labelfont=sf,textfont=sf]{subfig}
\usepackage{textcomp}
\usepackage{stfloats}
\usepackage{url}
\usepackage{verbatim}
\usepackage{graphicx}
\usepackage{graphics}
\usepackage{subfig}
\usepackage{caption}
\usepackage{cite}
\usepackage{tabularray}
\usepackage{hyperref}

\usepackage{xcolor}
\usepackage{ifthen}
\usepackage{etoolbox}

\usepackage[normalem]{ulem}

\newboolean{showchanges}

\setboolean{showchanges}{false} 

\ifthenelse{\boolean{showchanges}}{
    \usepackage{todonotes}
    \newcommand{\mynew}[1]{{\color{red}{#1}}}
    \newcommand{\myold}[1]{{\color{blue}\sout{#1}}}
}{
    \usepackage[disable]{todonotes}
    \newcommand{\mynew}[1]{{\color{black}{#1}}}
    \newcommand{\myold}[1]{\iffalse{#1}\fi}
}

\setuptodonotes{fancyline, color=blue!30, size=\tiny}

\begin{document}

\title{LiDAR-based Real-Time Object Detection and Tracking in Dynamic Environments}

\author{Wenqiang Du, Giovanni Beltrame, ~\IEEEmembership{Senior Member,~IEEE,}
\thanks{This paper was produced by MISTLab, the Department of Computer Engineering and
Software Engineering, Polytechnique Montreal. They are in Montreal, CA.}
\thanks{Manuscript received Month day, 2024; revised Month day, 2024.}}

\markboth{Journal of \LaTeX\ Class Files,~Vol.~14, No.~8, August~2021}%
{Shell \MakeLowercase{\textit{et al.}}: A Sample Article Using IEEEtran.cls for IEEE Journals}

\IEEEpubid{0000--0000/00\$00.00~\copyright~2021 IEEE}


\maketitle
\begin{abstract}
  In dynamic environments, the ability to detect and track moving objects in
  real-time is crucial for autonomous robots to navigate safely and effectively.
  Traditional methods for dynamic object detection rely on high accuracy
  odometry and maps to detect and track moving objects. However, these methods
  are not suitable for long-term operation in dynamic environments where the
  surrounding environment is constantly changing. In order to solve this
  problem, we propose a novel system for detecting and tracking dynamic objects
  in real-time using only LiDAR data.
  By emphasizing the extraction of low-frequency components from LiDAR data as
  feature points for foreground objects, our method significantly reduces the
  time required for object clustering and movement analysis. Additionally, we
  have developed a tracking approach that employs intensity-based
  ego-motion estimation along with a sliding window technique to assess object
  movements. This enables the precise identification of moving objects and
  enhances the system's resilience to odometry drift. Our experiments show that
  this system can detect and track dynamic objects in real-time with an average
  detection accuracy of 88.7\% and a recall rate of 89.1\%. Furthermore, our
  system demonstrates resilience against the prolonged drift typically
  associated with front-end only LiDAR odometry.

  All of the source code, labeled dataset, and the annotation tool are available at:
  \url{https://github.com/MISTLab/lidar_dynamic_objects_detection.git}
  

\end{abstract}

\begin{IEEEkeywords}
Point Cloud, LiDAR, Dynamic Object Detection, Ego-Motion Estimation, Real-Time, Autonomous Robotics
\end{IEEEkeywords}

\section{Introduction}
\IEEEPARstart{T}{ypically}, when estimating the movement of robots, the
assumption is that the surroundings are static\cite{ campos2021orb, zhang2014loam, qin2018vins}. However, this
assumption is now being challenged more frequently in recent research. This is
due to significant progress in the area of ego-motion estimation
\cite{xu2022fast, shan2021lvi, liosam2020shan} and an increasing need for
autonomous robots to work in dynamic environments\cite{furgale2013toward,
  peng2024review}, such as urban areas, warehouses, and construction sites. In
these settings, the presence of moving objects like pedestrians, cyclists, and
other vehicles can greatly affect the robot's ability to determine its location
and map the area. Consequently, the robot's capability to identify and track
moving objects in real time becomes a crucial task.

Detecting and tracking moving objects in real time is a complex task. It demands
that the robot can distinguish between static and moving objects, and track the
motion of these moving objects in real time. This task is particularly
challenging for robots that depend on Light Detection and Ranging (LiDAR) data
for perception, as the data is typically unordered and noisy. In contrast to
camera-based systems, which can obtain real-time semantic information of objects
for tracking \cite{redmon2016you, redmon2017yolo9000, redmon2018yolov3,
  bochkovskiy2020yolov4, ge2021yolox}, processing LiDAR data is not as
straightforward and efficient \mynew{due to the higher dimensionality, increased number of points, and the sparse nature of point clouds compared to image data}. \todo{why not efficient?}

\begin{figure}
  \centering
  \includegraphics[width=0.5\textwidth]{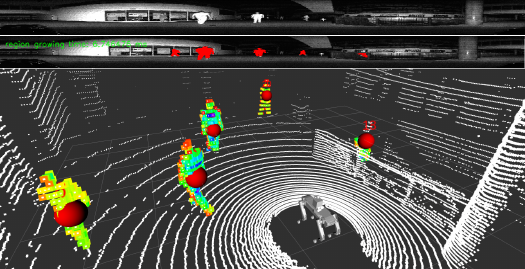}
  \caption{Illustration of dynamic object detection. Top: The raw intensity
    image. Middle: Dynamic objects are highlighted in red within the intensity
    image. Bottom: Dynamic points corresponding to these objects are extracted
    from the raw 3D point cloud, showing their spatial distribution.}
  \label{fig:dynamic_object_detection}
\end{figure}

\IEEEpubidadjcol At present, LiDAR-based techniques for identifying moving
objects are categorized into two primary groups: map-based methods, and
segmentation-based approaches\cite{lim2021erasor, arani2022comprehensive,
  schmid2023dynablox}, each coming with inherent limitations. Map-based methods
demand extensive computational resources for map creation, making them
unsuitable for real-time applications. Besides, those techniques struggle with
complex environments and require increased computational effort as the size of
the map grows. On the other hand, Segmentation-based approaches face challenges
in accurately detecting objects within complex environments filled with
unidentified objects, and their performance heavily depends on the objects'
visual characteristics.

To address these challenges, we propose a mapless dynamic object detection and
tracking system. Our method first transforms the unordered point cloud data into
an intensity image. This image then undergoes a process of 2D gaussian
convolution to reduce its high-frequency components. This step helps in reducing
noise and simplifying the distinction between foreground objects and the
background. The 3D points that belong to the features of these foreground
objects are retrieved from the original LiDAR data. By using intensity-based
ego-motion estimation\mynew{\cite{du2023real}} \todo{add citation}, our system can track
various object clusters in real-time. Analyzing the movement of these
clusters, the system can identify the central points of moving objects within
the environment and, through region growing, reconstruct the entire set of
points that make up these moving objects\cite{wu2024moving,kim2020remove}.

The main contributions of this work are summarized as follows:
\begin{enumerate}
\item \textbf{Innovative Processing of LiDAR Data:} We propose a novel method
  for processing LiDAR data by converting unordered point clouds into intensity
  images and applying 2D convolution. This technique effectively minimizes noise
  and accentuates the features of foreground objects, facilitating their
  identification and tracking. This represents a considerable advancement over
  conventional methods that struggle to process LiDAR data effectively;
\item \textbf{Efficient Object Clustering and Identification:} Employing
  intensity-based ego-motion estimation alongside a sliding window technique,
  our system excels in tracking clusters of objects in real time. Analyzing the
  movement of these clusters enables the identification of the central points of
  moving objects, leading to the accurate reconstruction of their full point
  sets. This method significantly enhances the system's efficiency in
  recognizing and tracking objects within its surroundings;
\item \textbf{Real-Time Detection and Tracking of Dynamic Objects:} Our method
  encompasses a robust system for detecting and tracking dynamic objects in real
  time. Using LiDAR data, we distinguish between static and dynamic objects and
  track their movements with high precision. This capability is essential for
  navigating environments populated with frequently moving objects;
\item \textbf{Robust Reconstruction of Dynamic Objects:} Using intensity
  image based region growing techniques, our method reconstructs the complete
  set of points that make up moving objects. This enables the identification
  of the objects' structures and positions within the environment,
  which is crucial for navigation and task execution. This robust reconstruction
  capability marks a significant improvement over prior techniques, which offer
  only partial or less precise representations.
\end{enumerate}

This article is structured as follows: Section \ref{sec:related_work} offers a
review of related work, focusing on the progress and challenges in
the dynamic object detection. This sets the stage for our research by providing
necessary background and justification. Section \ref{sec:methodology} elaborates
on our proposed methodology. It details the process of converting LiDAR point
clouds into intensity images, applying 2D convolution for background reduction,
and tracking dynamic objects using intensity-based ego-motion estimation,
clustering, estimation of movement combined with a sliding window approach.
Section \ref{sec:experimental_evaluation} is dedicated to the experimental
evaluation of our approach. It includes a comparative analysis with existing
methods, demonstrating the effectiveness of our system in handling dynamic
environments.
\myold{Section \ref{sec:discussion} addresses the limitations of our study and proposes
future research directions. The aim here is to suggest how further advancements
can be made in ego-motion estimation and dynamic object tracking. we discuss the
broader implications of our findings and the potential applications of our
method. This section also reflects on how our work contributes to the field of
robotics, particularly in terms of autonomous navigation. The article concludes
with Section \ref{sec:conclusion}, which summarizes the main contributions of
our work and highlights its significance in enhancing robotic capabilities for
navigating and operating in dynamic, complex environments.}

\mynew{Section \ref{sec:conclusion} summarizes the main contributions of
our work and highlights its significance in enhancing robotic capabilities for
navigating and operating in dynamic, complex environments. It also addresses the limitations of our study and proposes
future research directions, suggesting advancements in ego-motion estimation and dynamic object tracking.
We discuss the broader implications of our findings, the potential applications of our
method, and how our work contributes to the field of robotics, particularly in terms of autonomous navigation.}

\section{Related Work}
\label{sec:related_work}
\subsection{Map-based methods}
Map-based methods for the detection of dynamic objects can be categorized into
three primary types: ray-tracing methods, visibility-based methods, and
occupancy-based methods.

\paragraph*{Ray-Tracing Methods} 
Ray-tracing methodologies simulate LiDAR sensor behavior by tracing rays from
the LiDAR sensor outward into the environment\cite{azim2012detection,
  schauer2018peopleremover}. Rays are traced through either the grid map
\cite{thrun2003learning} or voxel map \cite{hornung2013octomap} to identify
dynamic objects. Should a ray traverse a grid cell or voxel, this indicates that
the cell is empty. If an object obstructs the ray within this empty cell, it
signifies that the object is dynamic.

Azim and Aycard \cite{azim2012detection} used an octree-based occupancy grid
map to represent the dynamic surroundings of the vehicle and identified moving
objects by noting discrepancies across scans. If the ray were reflected, then
that volume would be occupied. If the ray were to traverse the volume, then it
would be free. Peopleremover \cite{schauer2018peopleremover} efficiently removes
dynamic objects from 3D point cloud data by constructing a regular voxel
occupancy grid and traversing it along the sensor's line of sight to the
measured points to identify differences in voxel cells between different frames,
which correspond to dynamic points. It is applicable to both mobile mapping and
terrestrial scan data, aiming to produce a clean point cloud devoid of dynamic
objects, such as pedestrians and vehicles. This process conservatively removes
volumes only when they are confidently identified as dynamic, thus preserving
the integrity of static elements in the point cloud data. Khronos
\cite{schmid2024khronos} introduces a method for detecting dynamic objects in
robotic vision systems using ray tracing. Instead of relying on volumetric data,
it captures changes through surface representations and background and robot
poses. By creating a library of rays between observed background vertices and
robot positions, stored in a global hash map for efficiency, the method
dynamically detects objects without real-time free-space mapping. Object
detection is achieved by computing distances to rays, determining occlusion,
consistency, or absence. This technique offers a practical approach to real-time
object detection, balancing computational efficiency with accuracy. However, it
perform poorly in dynamic environments with large open spaces with sparse
surfaces. M-detector \cite{wu2024moving} is a motion event detection system
inspired by the human visual system's Magnocellular cells, known for their
motion sensitivity due to larger sizes and faster processing capabilities.
Using ray tracing and the principle of occlusion, it detects moving objects
by observing how they interrupt the LiDAR sensor's laser rays, effectively
occluding previously visible backgrounds or recursively occluding themselves.
The system processes inputs either as individual points or frames, applying
three parallel occlusion tests to determine motion events. This approach allows
for low-latency, high-accuracy detection without requiring extensive training
datasets. M-detector's design ensures it is highly generalizable, capable of
detecting various object sizes, colors, and shapes across different environments
with minimal computational resources, making it suitable for a wide range of
robotic applications.

However, Ray-tracing methods demonstrate a significant sensitivity to the
accuracy of odometry and incur substantial computational expenses during the
construction of maps and and as the map size gradually increases
\cite{lim2021erasor, lim2023erasor2, pomerleau2014long, banerjee2019lifelong}.
Additionally, these methods lead to increased computational times
\cite{schmid2024khronos} with prolonged operation and prove unsuitable for
real-time applications.

\subsubsection{Visibility-Based Methods}
Visibility-based algorithms operate on a fundamental physical premise: light
travels in straight lines, and if two points, one nearer and the other farther,
are detected on the same ray, the nearer point has to be
dynamic\cite{pomerleau2014long, ambrucs2014meta, jiang2016static} to allow
visibility of the farther point.
The core concept involves recognizing dynamic points by examining the geometric
variances between a query range image and its equivalent map range image. In
essence, if a pixel's range value in the query image surpasses the one in the
map image, it indicates an occlusion at the map image pixel's location according
to these methods. Therefore, points mapped onto these pixels are deemed
dynamic\cite{lim2023erasor2}.

Removert \cite{kim2020remove} introduces a novel method for the removal of
dynamic objects from LiDAR point cloud data. This method exploits the visibility
of points within the LiDAR data to identify dynamic objects. By comparing the
visibility of points in the current frame to that in the map, the method can
detect dynamic objects and subsequently remove them from the point cloud data.
This approach proves effective in removing dynamic objects from LiDAR data and
enhancing the accuracy of static object detection. However, it is an offline
method and is not suitable for real-time applications.
ReFusion \cite{palazzolo2019refusion} creates a dense mapping that can
effectively handle dynamic environmental elements without relying on specific
dynamic object class detection, using Truncated Signed Distance Function (TSDF)
and voxel hashing on GPU. The method detects dynamic objects by leveraging
residuals from the registration process and the explicit representation of free
space in the environment, achieving geometric filtering of dynamic elements.
However, maintaining a dense TSDF mapping can be computationally expensive and
memory-intensive.

The accuracy of the pose estimation is critical to the performance of the
visibility-based methods since they need to retrieve the corresponding points in
the map according to the odometry information. Moreover, false negatives, which
include but are not limited to self-interference within point clouds and false
positives caused by parallel point discrepancies, misjudged occluded points, and
contact point errors, directly undermine the reliability of dynamic points
extraction \cite{lim2021erasor}. Additionally, the invisibility of static
points, although less common than false negatives, poses a more challenging
problem to solve. For instance, when dynamic obstacles continuously block part
or all of the LiDAR's rays of light, the LiDAR fails to detect the static objects
behind these dynamic obstacles, leading to a scenario where the dynamic
obstacle's point cloud is never identified and filtered
out~\cite{lim2023erasor2}.

\subsubsection{Occupancy-Based Methods}
Occupancy-based methods employ statistical models to estimate the likelihood of
space being occupied, facilitating robust mapping of environments with dynamic
or uncertain elements.

Dynablox \cite{schmid2023dynablox} is a real-time, map-based dynamic object
detection system. It incrementally estimates high-confidence free-space areas
and leverages enhanced the Truncated Signed Distance Field (TSDF)
\cite{curless1996volumetric} method for volumetric mapping. This approach
enables robust detection of dynamic objects without assumptions about their
appearance or the structure of the environment. The method demonstrated an
Intersection over Union (IoU) of 86\% and processed data at 17 frames per second
(FPS) on a laptop-grade CPU in real-world datasets, surpassing appearance-based
classifiers and approximating the performance of certain offline
methods. \myold{ Equation (\ref{eq:iou}) has defined the IoU, maybe we can skip it here.} \todo{Define IoU. Maybe there's too much detail though...} However, the
study also acknowledges certain limitations, including challenges in detecting
extremely thin and sparsely measured objects, such as shades, due to the
voxel-based map representation. It also relies on prior odometry to place points
into map spaces. \myold{
ERASOR, developed by Lim et al.,
effectively removes dynamic objects from 3D point cloud maps using a novel
approach centered around the concept of Region-Wise Pseudo Occupancy Descriptor
(R-POD) and Region-wise Ground Plane Fitting (R-GPF). Pseudo occupancy is a type
of occupancy map that expresses the occupancy of unit space and then
discriminate spaces of varying occupancy. By focusing on the fundamental
observation that most dynamic objects in urban environments make contact with
the ground, ERASOR employs pseudo occupancy to evaluate the occupancy status of
space units, distinguishing between static and dynamic areas. The method then
utilizes R-GPF to accurately segregate static points from dynamic points within
these identified spaces. This process allows for the efficient removal of
dynamic objects from the point cloud data, ensuring a cleaner, more static map
for improved navigation and localization performance.
}
ERASOR2\cite{lim2023erasor2}\todo{Maybe just leave ERASOR2 and remove ERASOR} is
an instance level dynamic object removal system that can effectively remove
dynamic points from 3D point cloud maps. It uses Pseudo Occupancy Grid Map to
represent whether the regions are temporally occupied or not by calculating the
probability of those regions. It also employs instance segmentation
\cite{nunes2022unsupervised} estimates to identify and exclude dynamic points at
the instance level, ensuring that static points are preserved with minimal loss.
This approach allows for the precise removal of dynamic objects from the map,
enhancing the map's utility for navigation and planning tasks.

Nevertheless, these methods still require maintaining a map of the environment,
which is memory-intensive and computationally demanding, especially for
long runs.

\begin{figure*}[h]
  \centering
  \includegraphics[width=1\textwidth]{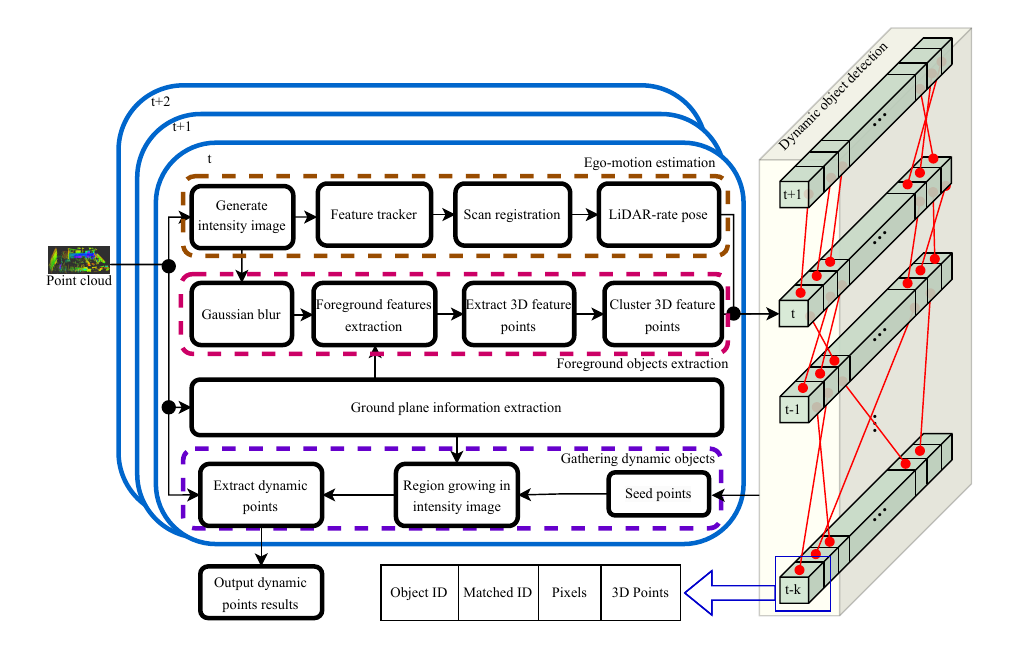}
  \caption{The framework of our dynamic object detection system, consisting of
    three main components: data preprocessing, object tracking, and dynamic
    object detection. The data preprocessing component converts unordered LiDAR
    point cloud data into intensity images and extracts low-frequency components
    to represent foreground objects. Object tracking involves ego-motion
    estimation, clustering, and cluster association. The ego-motion estimation
    component calculates the robot's movement between frames using
    intensity-based front-end odometry, providing real-time movement data
    despite potential noise and drift. The dynamic object detection component
    leverages data association results to track objects and identify seed points
    of dynamic objects, followed by a region growing algorithm to reconstruct
    the complete point set of dynamic objects from these seed points. This
    system processes LiDAR data in real-time to detect dynamic objects in
    complex environments, using the low-frequency components to represent
    foreground objects and enable real-time tracking across multiple frames.}
  \label{fig:framework}
  
\end{figure*}

\subsection{Segmentation-Based Methods}
Segmentation-based methods focus on identifying dynamic objects by semantically
segmenting the point cloud and analyzing the movement of points or objects over
time. These methods are particularly effective in detecting dynamic objects in
complex environments with unidentified objects.

Thomas et al. \cite{thomas2022learning} proposed a self-supervised learning
method for dynamic object detection in LiDAR point clouds. They first
projected each frame of point cloud into 2D grids. Then, they stacked the 2D
frames into 3D according to the time sequence. Finally, they used a KPConv
network \cite{thomas2019kpconv} 3D back-end and a three levels U-Net
\cite{ronneberger2015u} 2D front-end to predict the future time steps of
spatiotemporal occupancy grid maps. Chen et al. \cite{chen2021moving} used
the range image to represent the point cloud data and combined the current frame
and residual images generated between current frame and the last few frames.
After then, they used a range projection-based segmentation CNNs
\cite{milioto2019rangenet++, li2021multi, cortinhal2020salsanext} to predict the
moving objects.

As an upgrade to \cite{thomas2022learning}, Sun et al. \cite{sun2022efficient}
extended the SalsaNext \cite{cortinhal2020salsanext} by adopting a dual-branch (a
range image branch and a residual image branch) and dual-head (an image head and
a point head) architecture to separately deal with spatial and temporal
information. After then, they used a 3D sparse convolution to refine the
prediction results. In this coarse-to-fine framework, the network can predict
the moving objects more accurately and efficiently. Mersch et al.~\cite{mersch2022receding}
combined a sequence of LiDAR point cloud into a voxelized spare 4D point cloud.
Then, they feeded the combined 4D occupancy grid to a modified MinkUNet14
\cite{choy20194d} to extract spatio-temporal features and predicted the
confidence score for each point.

These methods are effective in detecting dynamic objects; however, they
predominantly depend on a substantial volume of annotations for the dynamic
objects in the training data, which may not always be accessible. Furthermore,
segmentation-based methods incur significant computational costs and frequently
yield unsatisfactory results in unfamiliar environments. Consequently, a need
persists for a lightweight and label-free method for dynamic object detection in
real-time.

In this work, to increase speed and reduce computational load, we
primarily concentrate on the low-frequency components of LiDAR data, which are
used to represent foreground objects. Given that the size of the
low-frequency components is significantly smaller than that of the original
LiDAR data, our approach enables real-time tracking of multiple objects across
multiple frames.

\section{Methodology}
\label{sec:methodology}
Our proposed method is illustrated in Fig \ref{fig:framework}. This framework
principally comprises three components: data preprocessing, object tracking, and
dynamic object detection. Our method is engineered to process

\subsection{Data Preprocessing}
\label{sec:data_preprocessing}
\subsubsection{Point Cloud to Intensity Image}
\label{sec:point_cloud_to_intensity_image}

Unlike cameras, which generate an image in each cycle, LiDARs produce a frame of
point cloud data per cycle. Each point $\mathbf{p}$ within this frame records
spatial dimensions and intensity information, $\mathbf{p} = \{x, y, z, i\}$. The
intensity of a point is determined by the signal, which constitutes the number
of photons collected by the sensor for a pixel during a single cycle. The
initial step of our method involves converting the unordered LiDAR point cloud
data into an intensity image through spherical projection. This process entails
projecting the point cloud data onto a two-dimensional grid; each grid cell
therein represents the intensity of the points contained within it. The grid
cell's index $(u, v)$ can be determined by the following equation:


\begin{equation}
  \label{eq:spherical_projection}
  \left[
    \begin{array}{c}
      u\\  
      v
    \end{array}
   \right]
  = \left[
    \begin{array}{c}
      \frac{\left( \arctan(y, x) + \pi \right)}{2\pi} \cdot w \\  
      \frac{\left( \beta_{up} - \arcsin(zr^{-1}) \right)}{\beta_{fov}} \cdot h
    \end{array}
  \right]
\end{equation}
Where the $r$ represents the distance between the point and the sensor, denoted
as $||\mathbf{p}||_2$, $w$ and $h$ are the width and height of the image,
respectively, and $\beta_{up}$ and $\beta_{fov}$ are the upper bound and field
of view of the LiDAR sensor, respectively.

By converting the point cloud data into an intensity image, we treat the
unordered LiDAR data similarly to an image, simplifying subsequent processing
steps. This transformation permits the application of image processing
techniques to LiDAR data, thus facilitating the identification and tracking of
dynamic objects.
\subsubsection{Low-frequency components extraction} 
\label{sec:lower_frequency_components_extraction}
The intensity image generated in the previous step contains high-frequency
components that can significantly slow down the clustering and tracking
processes. In order to extract features of the foreground objects
with fewer points, we apply a 2D gaussian convolution to the intensity image to
extract low-frenquency components, see (\ref{eq:gaussian_convolution}), and
treat them as a representative of foreground objects. This process involves
convolving the intensity image with a 2D Gaussian kernel which effectively
smooths the image and minimizes noise:

\begin{equation}
  \label{eq:gaussian_kernel}
  g(m, n) = \frac{1}{2\pi \sigma_m \sigma_n} \exp\left(-\left(\frac{m^2}{2\sigma_m^2} + \frac{n^2}{2\sigma_n^2}\right)\right)
\end{equation}
Here, $m$ and $n$ represent the distances from the center of the kernel to the
current point and can take integer values, including negative ones. $m$ ranges
from $[-a, a]$ and $n$ ranges from $[-b, b]$. In this case, as the intensity
image is quite flat, we also configure the kernel to have a flat shape, where $a
> b$. This configuration ensures that the Gaussian kernel matrix's index for the
center of columns is 0, with $a$ positions on either side, and similarly, the
center index for rows is 0, with $b$ positions on each side. $\sigma_m$ and
$\sigma_n$ denote the standard deviations along the horizontal and vertical
directions, respectively. The Gaussian kernel size is $(2a+1, 2b+1)$. Then, we
convolve the intensity image with the Gaussian kernel to extract the
low-frequency components:

\begin{equation}
  \label{eq:gaussian_convolution}
  \begin{aligned}
    h(u, v) &= (f * g)(u, v) \\
            &= \sum_{m=-a}^{a} \sum_{n=-b}^{b} f(u-m, v-n) \cdot g(m, n) \\
  \end{aligned}
\end{equation}
where $f(u-m, v-n)$ represents the intensity value of the image at
position $(u-m, v-n)$, and $g(m, n)$ represents the value of the kernel at
position $(m, n)$. The variables $m$ and $n$ are the distances from the center
of the kernel, ranging over the kernel dimensions. The convolution operation is
performed by sliding the kernel over the image and calculating the sum of the
products of the kernel values and the image values at each position.

Finally, we subtract the blurred image from the original intensity image, as
shown in (\ref{eq:feature_image}). The resulting image retains only the
low-frequency components, which depict the features of the foreground objects.
By focusing on these low-frequency components, we are able to accurately
identify and track dynamic objects in real-time.

\begin{equation}
  \label{eq:feature_image}
  t(u, v) = B\left(f(u, v) - h(u, v), \theta\right)
\end{equation}
where $ B(x, \theta) $ is a binarization function defined by:
\begin{equation*}
  B(x, \theta) = 
  \begin{cases} 
  1 & \text{if } x > \theta \\
  0 & \text{otherwise}
  \end{cases}
\end{equation*}
and $ \theta $ is the threshold value.

\subsection{Object Tracking}
After extracting the low-frequency components from the intensity image, the 3D
points associated with the features of these foreground objects can be further
extracted from the raw LiDAR data. Subsequently, these points can be clustered
into distinct objects and tracked in real-time, owing to their small number.

\subsubsection{Intensity-based Ego-Motion Estimation}
The initial step in object tracking involves estimating the ego-motion of the
robot. This is accomplished by extracting ORB features~\cite{rublee2011orb} from
the intensity image and calculating the transformation matrix between
consecutive frames. The resulting transformation matrix is used to estimate the
robot's movement between frames, an essential factor for accurate object
tracking. This technique, known as intensity-based ego-motion estimation, was
proposed in our previous work~\cite{du2023real}. Here we only use the front-end
of the SLAM system to make is lightweight and real-time, but drift may occur
during long-term operation. However, in the following we show that
drift will not affect the object tracking accuracy.

For each frame of intensity image, we extract ORB features and match them with
the features in the previous frame. At the same time, we can extract the
corresponding 3D points from current and previous frames and put them into
$\mathcal{X} = \{\mathbf{X}_1, \mathbf{X}_2, \cdots, \mathbf{X}_k \}$ and
$\mathcal{Y} = \{\mathbf{Y}_1, \mathbf{Y}_2, \cdots, \mathbf{Y}_k \}$,
respectively. Then, the scan match problem can be formulated as a least square
estimation problem:

\begin{equation}
  \label{eq:argmin}
  \mathop{\arg\min}_{\mathbf{R,t}} \sum\limits_{n\in \mathcal{N}} \parallel \mathbf{Y}_n - (\mathbf{R}_i\mathbf{X}_n + \mathbf{t}_i)\parallel_2
\end{equation}
where $\mathcal{N}=[1, 2, \cdots, k]$, $\mathbf{X}_n = \{ x_n, y_n, z_n\}$, and
$\mathbf{R}_i$ and $\mathbf{t}_i$ are the rotation matrix and translation vector
at current frame time $i$ with respect to the previous frames, respectively. The
solution of this optimization problem is the ego-motion of the robot between two
frames. So, we need to accumulate $\mathbf{R}_i$ and $\mathbf{t}_i$ to get the
robot's pose in a reference coordinate system.

\subsubsection{Clustering}
Since the number of objects varies across different frames, traditional
clustering methods that require predefined cluster numbers, such as K-means,
cannot be employed. Furthermore, the vertical disparity between two adjacent
laser beams increases as the distance to the objects increases. Consequently,
addressing this challenge demands the use of a more robust clustering
method. We initially employ an Euclidean distance-based clustering
method to segment all points into groups, which are subsequently clustered into
distinct objects.

As mentioned above, we can get 3D
points $\mathcal{Q} = \{\mathbf{p}_0, \mathbf{p}_1, \cdots, \mathbf{p}_n\}$
that represent the features of the foreground objects. For each point in
$\mathcal{Q}$, we calculate the Euclidean distance between it and all other
points in horizontal and vertical plane, see (\ref{eq:euclidean_distance}).
\begin{equation}
  \label{eq:euclidean_distance}
  \mathbf{d}({\mathbf{p}_i}, \mathbf{p}_j)=
  \left[
    \begin{array}{c}
    
      d_{xy}(\mathbf{p}_i, \mathbf{p}_j)\\
      d_z(\mathbf{p}_i, \mathbf{p}_j)
    \end{array}
  \right]
  =
  \left[
    \begin{array}{c}
      \sqrt{(x_i - x_j)^2 + (y_i - y_j)^2 }\\
      |z_i - z_j|
    \end{array}
  \right]  
\end{equation}
The points $i$ and $j$ can be clustered into the same group $\mathcal{C}_m =
\{\mathbf{p}_i, \mathbf{p}_j, \cdots, \mathbf{p}_k\}$ if the distance between
them is less than a predefined threshold, $\epsilon$. The clustering process
can be formulated as (\ref{eq:clustering}):
\begin{equation}
  \label{eq:clustering}
    \forall \mathbf{p}_i, \mathbf{p}_j \in \mathcal{C}_m, \quad \mathbf{d}({\mathbf{p}_i}, \mathbf{p}_j) \leq \mathbf{\epsilon}
\end{equation}
After clustering, we get a group of clusters $C = \{\mathcal{C}_1,
\mathcal{C}_2, \cdots, \mathcal{C}_m, \cdots, \mathcal{C}_q\}$, and the total
number of cluster $q$ can be different for each frame.

\subsubsection{Cluster Association}
The clusters need to be linked between frames to ensure the continuity of the
tracking process. We use the Global
Nearest Neighbor (GNN)~\cite{bar1995multitarget} method for this association.
We first calculate the center point of each cluster in the current frame,
$\mathbf{c}_m^{i} = [x_m, y_m, z_m]^T$, and the center point of each cluster in
the previous frame, $\mathbf{c}_n^{i-1} = [x_n, y_n, z_n]^T$. Then, we calculate
the Euclidean distance between the center points of the clusters in the current
frame and the previous frame:

\begin{equation}
  \label{eq:cluster_association}
  d_{\mathbf{c}}(\mathbf{c}_m^{i}, \mathbf{c}_n^{i-1}) = \parallel \mathbf{c}_m^{i} - (\mathbf{R}_i\mathbf{c}_n^{i-1} + \mathbf{t}_i)\parallel_2
\end{equation}
Subsequently, we will build a cost matrix $\mathbf{D} \in \mathbb{R}^{M \times
  N}$, where $M$ and $N$ are the number of clusters in the current frame and the
previous frame, respectively. Each element $d_{m, n} =
d_{\mathbf{c}}(\mathbf{c}_m^{i}, \mathbf{c}_n^{i-1})$ in the cost matrix
represent the distance cost if the cluster $m$ in the current frame matches with the
cluster $n$ in the previous frame. To filter out the unmatched clusters,
distances exceeding the threshold $d_{max}$ are assigned a prohibitively high
cost to prevent them from being selected in the optimization process:
\begin{equation}
  \label{eq:cost_matrix}
  d_{m,n} = 
  \begin{cases} 
    d_{\mathbf{c}}(\mathbf{c}_m^{i}, \mathbf{c}_n^{i-1}) & \text{if } d_{\mathbf{c}}(\mathbf{c}_m^{i}, \mathbf{c}_n^{i-1}) \leq d_{\text{max}} \\
  \infty & \text{otherwise}
  \end{cases}
\end{equation}
It is possible that some previously appearing objects do not appear in the
current frame, while new objects might appear. To solve this problem, we add
dummy targets to the cost matrix. The cost between dummy targets and real
objects is set to a high value, which is $2d_{max}$ as penalty costs while the
cost between dummy targets is set to 0. This way, the cost matrix can be
extended to $\mathbf{D} \in \mathbb{R}^{(M+M') \times (N+N')}$. We can then use
the Hungarian algorithm \cite{kuhn1955hungarian} to solve the assignment problem
and get the optimal solution\myold{, see(\ref{eq:gaussian_convolution})}. The Hungarian
algorithm is a combinatorial optimization algorithm that solves the assignment
problem in polynomial time,
ensuring that clusters are tracked across frames:

\begin{equation}
  \min \sum_{m=1}^{M+M'} \sum_{n=1}^{N+N'} \alpha_{mn} d_{m,n}
  \label{eq:optimization_modified}
\end{equation}
where:
$ M' $ and $ N' $ represent the number of dummy targets added to the current and previous frames respectively.
Subject to the constraints:
\begin{equation}
  \sum_{m=1}^{M+M'} \alpha_{mn} = 1 \quad \text{for all } n \in \{1, \ldots, M+M'\}
  \label{eq:constraint1_modified}
\end{equation}
\begin{equation}
  \sum_{n=1}^{N+N'} \alpha_{mn} = 1 \quad \text{for all } m \in \{1, \ldots, N+N'\}
  \label{eq:constraint2_modified}
\end{equation}

\begin{equation}
  \alpha_{mn} \in \{0, 1\}
  \label{eq:binary_constraint_modified}
\end{equation}
By employing this method, we guarantee that each cluster in the current frame is
matched uniquely with a corresponding cluster from the previous frame, thus
preserving the continuity of the tracking process.



\subsection{Dynamic Object Detection}
\subsubsection{Movement Analysis}
Following the process of cluster association, we can determine all matching
clusters for all frames across a given time slot. To analyse the movement of the
clusters in current frame, we first create a sliding window of a fixed size, $k$
from the initial to the current
frame. 

We accumulate the transformation matrix $\mathbf{T}_i$ between frame $i$ and the
initial frame. From (\ref{eq:argmin}), we can get the relative rotation
$\mathbf{R}_i$ and translation matrix $\mathbf{t}_i$ between current frame and
previous frame. To simplify the following expression, we transform them from SO3
to SE3 and perform a homogeneous transformation on the cluster's centroid.
\begin{equation}
  \label{eq:transformation_matrix}
  \mathbf{T} = \left[\begin{array}{cc}\mathbf{R}& \mathbf{t} \\ \mathbf{0}^T& 1 \end{array} \right]= \left[\begin{array}{cccc}rx_x & ry_x & rz_x & x \\ rx_y & ry_y & rz_y & y\\ rx_z & ry_z & rz_z & z\\ 0&0&0& 1\end{array} \right]
\end{equation}
\begin{equation}
  \label{eq:homogeneous_transformation}
  \mathbf{c}_m^{i} = [x_m^i, y_m^i, z_m^i, 1]^T
\end{equation}
For frames inside the sliding window, we can get a series of transformation
matrices $\mathbf{T}_i$, $i \in \{1, 2, \cdots, k\}$. Then, we can calculate the
relative transformation matrix between the frame $i$ and the initial frame:

\begin{equation}
  \label{eq:relative_transformation}
  \mathbf{T}_i^w = \prod_{j=1}^{i} \mathbf{T}_j
\end{equation}
Where, $\mathbf{T}_i^w$ is the relative transformation matrix between the frame
$i$ and the initial frame, and $\mathbf{T}_j$ is the transformation matrix
between the frame $j$ and frame $j-1$.

Then, we can transform the cluster's centroid in the current frame to the
initial frame:
\begin{equation}
  \label{eq:transformation}
  \mathbf{c}_m^{wi} = \mathbf{T}_i^{w} \mathbf{c}_m^{i}
\end{equation}
Here, $\mathbf{c}_m^{wi}$ represents the centroid of cluster $m$ in the $i$th
frame within the initial frame's coordinate system. Having already transformed
the centroid of the cluster from the current frame to the initial frame, we can
now employ the format used prior to the homogeneous transformation.
Consequently, the cluster's centroid in the initial frame can be represented as:
\begin{equation}
  \label{eq:before_homogeneous_transformation}
  \mathbf{c}_m^{wi} = [x_m^{wi}, y_m^{wi}, z_m^{wi}]^T
\end{equation}
\mynew{where $x_m^{wi}$, $y_m^{wi}$, and $z_m^{wi}$ represent the $x$, $y$, and $z$ coordinates of the centroid position of cluster $m$ of the $i$th frame in the initial frame's coordinate system.}
\todo[inline]{I think we are missing the description of some terms}

\begin{figure}
  \centering
  \includegraphics[width=0.5\textwidth]{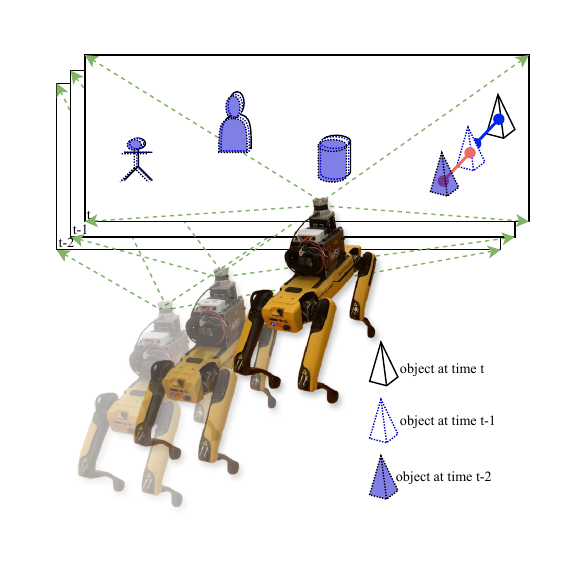}
  \caption{Dynamic Object Tracking: to adapt to the drift of the front-end
    odometry, we accumulate the transformation matrix between the current frame
    and the first frame within a time window. This approach allows us to ignore
    historical drift and only account for the drift within the current time
    window. By transforming all frames into the same coordinate system, we can
    calculate the distance between the centroids of clusters in consecutive
    frames and match corresponding clusters between frames. This method enables
    real-time tracking of all detected foreground objects. By analyzing the
    movement of clusters within the sliding window, we can effectively detect
    dynamic objects. In the picture, three frames within the time window are
    transformed into the initial frame's coordinate system. Although all
    clusters appear to move due to the odometry drift, the movement of static
    objects is minimal and usually follows a zero-mean Gaussian distribution.
    Therefore, we can filter out dynamic objects by accumulating the distance
    between the cluster's centroid in the current frame and the initial frame.}
  \label{fig:object_tracking}
\end{figure}

We can then track pairs of matched clusters to the initial frame if they exist.
Figure \ref{fig:object_tracking} shows the object tracking
principles.\todo{Maybe the principles should come before the details?}
We can finally calculate the Euclidean distance between the cluster's
centroid in the current frame and the initial frame:
\begin{equation}
  \label{eq:distance}
  f(\mathbf{c}_m^{wk}, \mathbf{c}_m^{w1}) = \parallel \mathbf{c}_m^{wk} - \mathbf{c}_m^{w1}\parallel_2
\end{equation}
if the distance is less than a predefined threshold, $\epsilon_d$, we can
consider the cluster as a dynamic object.

\subsubsection{Outlier Detection}
As we only rely on the front-end odometry, its drift may cause some outliers to
appear as dynamic objects. To solve this
problem, we can calculate the absolute distance of the matched clusters in the
current frame:
\begin{equation}
  \label{eq:outlier_1}
  f_a(\mathbf{c}_m^{wk}, \mathbf{c}_m^{1}) = \sum_{j=2}^{k} \parallel \mathbf{c}_m^{wj} - \mathbf{c}_m^{w(j-1)}\parallel_2
\end{equation}
if the object is static and the drift noise follows a Gaussian distribution,
$f(\mathbf{c}_m^{wk}, \mathbf{c}_m^{w1}) \approx 0$ and $f_a(\mathbf{c}_m^{wk},
\mathbf{c}_m^{1}) > 0$. So, we can use the ratio of $f(\mathbf{c}_m^{wk},
\mathbf{c}_m^{w1})$ and $f_a(\mathbf{c}_m^{wk}, \mathbf{c}_m^{1})$ to filter out
the outliers. If the ratio is less than a predefined threshold, $\epsilon_o$, we
can consider the cluster as a dynamic object.

In addition to the drifting problem,\myold{ we also encounter the new arrival problem}
\mynew{we are also facing an issue with long objects being mistakenly identified as dynamic objects due to the sparse nature of LiDAR. As the robot moves forward and the LiDAR scans the surface of such objects, the shape of the cluster remains almost the same, causing the center of the object to appear to shift, creating the illusion of movement.}
\todo{Maybe give an intuition of what it is? Why is it called this way?}
\myold{A cluster tends to grow as the robot moves forward, especially for an object
which is long and parallel to the direction of movement.}\mynew{Such clusters tend to grow as the robot moves forward, especially for objects that are long and parallel to the direction of movement.} To solve this problem,
we propose to calculate the cluster's moving direction and the cluster's main
direction and compare the angle between them. If the angle is less than a
predefined threshold, $\epsilon_{\theta}$, we can consider the cluster as a
dynamic object. For a cluster $m$ in the current frame, we can calculate the
moving direction as:
\begin{equation}
  \label{eq:moving_direction}
  \vec{v} = \frac{\mathbf{c}_m^{wk} - \mathbf{c}_m^{w1}}{\parallel \mathbf{c}_m^{wk} - \mathbf{c}_m^{w1}\parallel_2}.
\end{equation}

For the cluster $\mathcal{C}_m$, we can use Principal
Component Analysis (PCA) to calculate its main direction $\mathbf{V}$:
\begin{equation}
  \label{eq:pca}
  \mathbf{V} = \text{PCA}(\mathcal{C}_m).
\end{equation}
The angle between the moving direction and the main direction is then
\begin{equation}
  \label{eq:angle}
  \theta = \arccos(\vec{v} \cdot \mathbf{V}).
\end{equation}
If $\theta$ is smaller than a predefined threshold, $\epsilon_{\theta}$, we can
consider the cluster as an outlier.

\subsubsection{Region Growing}
After removing the outliers, we have seed points for the dynamic objects in
the current frame. After converting the seeds points from the initial
coordinate frame to the current, we can get the
dynamic objects' center points's pixel position in the raw intensity image (see
Figure~\ref{fig:dynamic_object_detection}, top). A region growing method can be
sued on the seed points to segment the whole dynamic object.
The basic idea is to start from the seed points and grow the region by adding
the neighboring points that have similar properties to the seed points. We use
the Euclidean distance between the points to determine whether they are similar.
However, finding the neighboring points in the raw point cloud is
time-consuming. To solve this problem, we can use the intensity image to find
the neighboring points. For each point in the intensity image, we can extract
the surrounding pixels directly as neighboring points instead of comparing their
Euclidean distance in the raw point cloud.
The neighbor points of each seed points' pixel index can be extracted as:
\begin{equation}
  \label{eq:neighbor_points}
  N_8(x, y) = \{(x+i, y+j) \mid i, j \in \{-1, 0, 1\}, (i, j) \neq (0, 0)\}
\end{equation}
Where, $N_8(x, y)$ is the 8-neighborhood of the seeds point's pixel $(x, y)$ in
the intensity image.

The region growing process can be formulated as
\begin{equation}
  \label{eq:region_growing}
  \mathcal{Q}_{\text{new}} = \mathcal{Q} \cup \left\{ \mathbf{p}_j \mid j \in N_8(i_x, i_y), \exists \mathbf{p}_i \in \mathcal{Q},  \quad \parallel \mathbf{p}_i, \mathbf{p}_j \parallel_2 \leq \mathbf{\epsilon} \right\}
\end{equation}
where, $\mathcal{Q}$ is the set of seed points of the dynamic object, and
$\mathbf{p}_j$ is a neighboring point of the seed points. If the distance
between the neighboring points and a seed point is less than a predefined
threshold, $\epsilon$, we can add the neighboring points to the dynamic object.
This process is repeated until all eligible neighboring points are added to the
dynamic object.

To prevent regions from absorbing all points we
use the ground plane information to filter out neighboring points that do
not belong to the dynamic object. Figure~\ref{fig:dynamic_object_detection},
middle, shows the dynamic object after the region growing process in red.
Figure \ref{fig:dynamic_object_detection}, bottom, shows the dynamic object
points retrieved from the raw point cloud according to the dynamic object's pixel
index in the intensity image.

\begin{figure}
  \centering
  \includegraphics[width=0.48\textwidth]{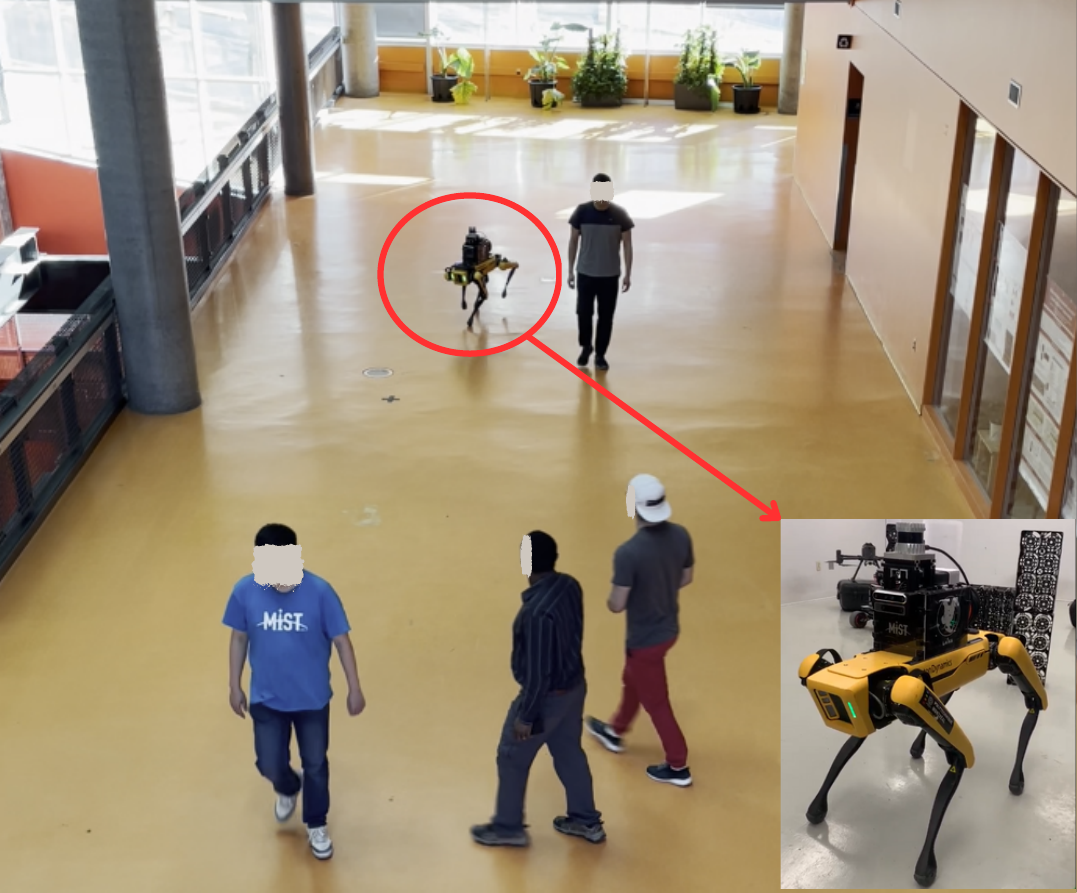}
  \caption{Spot robot navigating a dynamic environment. Equipped with an OS0-64
    LiDAR sensor, the Spot robot moves through a large area filled with
    pedestrians as moving objects. Both the pedestrians and the Spot robot
    exhibit random movements, creating a constantly changing environment that
    challenges the robot's navigation and object detection capabilities.}
  \label{fig:scenario}
\end{figure}

\section{Experimental Evaluation}
\label{sec:experimental_evaluation}
\subsection{Dataset}
\begin{figure}[hb!]
  \centering
  \includegraphics[width=0.48\textwidth]{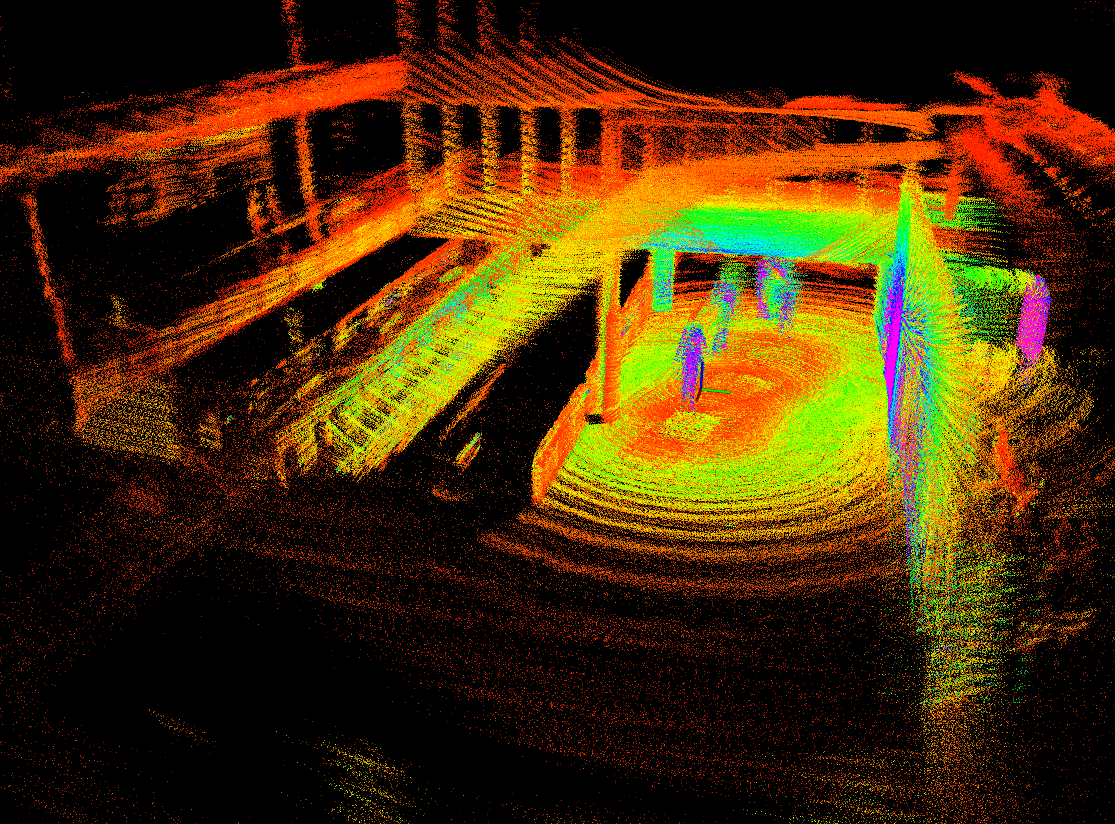}
  \caption{Point cloud map of the environment with dynamic objects. The dynamic points create blurring in the map as they move, highlighting the challenges of accurately mapping in dynamic environments.}
  \label{fig:map_with_dynamic_points}
\end{figure}

We used a Boston Dynamic Spot robot equipped with an Ouster 64-line LiDAR sensor
to collect the dataset. The robot was driven around the floor of a large space
with moving objects, shown in Figure \ref{fig:scenario}. In normal conditions,
the map is blurred by the movement of the pedestrians
(Figure~\ref{fig:map_with_dynamic_points}), and the moving objects affect the
localization accuracy. We manually labeled the dynamic objects in the dataset to
evaluate the performance of the proposed method. The labeled information was
stored in the following format: \{timestamp, dynamic points index\}. The
timestamp represents the time when the dynamic objects were detected, and the
dynamic points index represents the pixel index of the dynamic objects in the
intensity image.

To efficiently label streaming 3D dynamic points, we developed a specialized
tool that allows users to label dynamic objects within an intensity image in a
streaming style and store the labeled information in \myold{text}\mynew{binary}\todo{or binary}
format.

When labeling dynamic objects, users can check previous and following frames to
identify which points are dynamic. Users can then click the center of the
dynamic object in the intensity image and apply a region growing method to
capture all points belonging to the dynamic object. To aid in this process, the
tool provides visual assistance by displaying the extracted foreground points in
the intensity image, along with ground information.
The labeling tool will be open-sourced\footnote{\url{https://github.com/MISTLab/lidar_dynamic_objects_detection.git}}

\subsection{Evaluation Metrics}
\begin{figure}[hb!]
  \centering
  \includegraphics[width=0.48\textwidth]{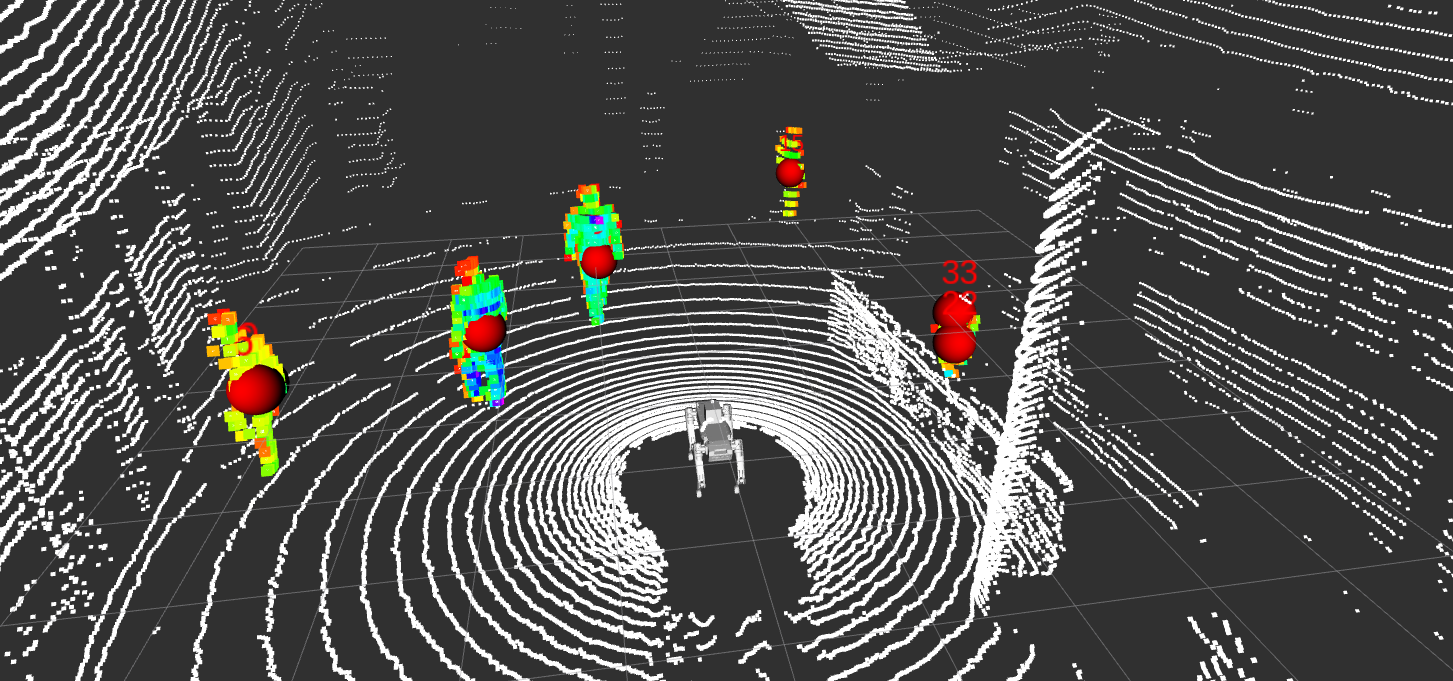}
  \caption{Segmentation results of dynamic objects in a single frame. The dynamic objects are segmented from the raw point cloud and displayed with colorful points. The center points of the dynamic objects are marked in red.}
  \label{fig:dynamic_object_points}
\end{figure}

Due to the time consuming nature of labelling, we
only labeled four sequences of the data collected from the Spot robot to assess
the performance of the proposed method. The estimated dynamic objects are
recorded in the same format as the labeled data. The performance of the proposed
method is evaluated by comparing the labeled data with the ground truth provided
by labeling. The estimated dynamic objects are showing in figure
\ref{fig:dynamic_object_points} with colorful points.

We used the following metrics to evaluate the performance of the proposed
method: precision, recall, \myold{intersection over union (IoU)}\mynew{IoU}, and F1 score\mynew{\cite{wu2024moving,lim2023erasor2,goutte2005probabilistic}}\todo{Reference!}. 
These
metrics are commonly used in object detection and tracking tasks and provide a
comprehensive evaluation of the method's performance:
\begin{itemize}
  \item \textbf{Precision}: The ratio of correctly identified positive instances to the total instances identified as positive.
  \begin{equation}
      \text{Precision} = \frac{TP}{TP + FP}
  \end{equation}
  \item \textbf{Recall}: The ratio of correctly identified positive instances to the total actual positive instances.
  \begin{equation}
      \text{Recall} = \frac{TP}{TP + FN}
  \end{equation}
  \item \textbf{IoU}: The ratio of the overlap between the predicted and ground truth bounding boxes to the union of these boxes.
  \begin{equation}
      \label{eq:iou}
      \text{IoU} = \frac{\text{Area of Overlap}}{\text{Area of Union}} = \frac{TP}{FN+TP+FP}
  \end{equation}
  \item \textbf{F1 Score}: The harmonic mean of precision and recall, providing a single metric that balances both aspects.
  \begin{equation}
      \text{F1 Score} = \frac{2 \cdot \text{Precision} \cdot \text{Recall}}{\text{Precision} + \text{Recall}}
  \end{equation}
\end{itemize}
Where $TP$ is the number of true positive points, $FP$ is the number of false
positive points, and $FN$ is the number of false negative points. The precision,
recall, IoU, and F1 score are calculated based on the number of true positive,
false positive, and false negative points, providing a comprehensive evaluation
of the method's performance.

\begin{table*}[ht!]
  \centering
  \caption{Performance metrics comparison across different methods. The best results are highlighted in bold, and the second-best results are underlined. The table compares the precision, IoU (Intersection over Union), recall, and F1 score for M-detector with front-end only odometry, Removert with front-end only odometry, M-detector with FAST-LIO, Removert with A-LOAM, and our proposed method with front-end only odometry across four sequences.}
  \begin{tblr}{
    colspec = {Q[2cm] Q[5cm] Q[2cm] Q[2cm] Q[2cm] Q[2cm]},
    row{1} = {font=\bfseries},
    cell{even}{2-6} = {gray9},
  }
  \hline
  Sequence    & Method                                  & Precision & IoU  & Recall & F1 Score \\
  \hline
              & M-detector w/ front-end only Odom       & 0.471     & {0.455} & {0.837}   & {0.578}     \\
              & Removert w/ front-end only Odom         & 0.080     & 0.063 & 0.287  & 0.117     \\
  Sequence 1  & M-detector w/ FAST-LIO                  & \underline{0.695} & \underline{0.669} & \textbf{0.915} & \underline{0.775} \\
              & Removert w/ A-LOAM                      & 0.651     & 0.460 & 0.616   & 0.610     \\
              & Ours w/ front-end only Odom             & \textbf{0.871} & \textbf{0.842} & \underline{0.898}   & \textbf{0.875}     \\
  \hline
              & M-detector w/ front-end only Odom       & 0.486     & {0.452} & {0.751}   & {0.547}     \\
              & Removert w/ front-end only Odom         & 0.031     & 0.024 & 0.141   & 0.046     \\
  Sequence 2  & M-detector w/ FAST-LIO                  & \underline{0.775} & \underline{0.677} & \underline{0.800}   & \underline{0.759}     \\
              & Removert w/ A-LOAM                      & {0.766} & 0.366 & 0.421   & 0.524     \\
              & Ours w/ front-end only Odom             & \textbf{0.883}     & \textbf{0.876} & \textbf{0.899}   & \textbf{0.888}     \\
  \hline
              & M-detector w/ front-end only Odom       & 0.642     & {0.588} & \underline{0.871}   & {0.703}     \\
              & Removert w/ front-end only Odom         & 0.149     & 0.072 & 0.200   & 0.134     \\
  Sequence 3  & M-detector w/ FAST-LIO                  & \underline{0.801} & \underline{0.728} & 0.844   & \underline{0.808}     \\
              & Removert w/ A-LOAM                      & 0.787     & 0.376 & 0.421   & 0.527     \\
              & Ours w/ front-end only Odom             & \textbf{0.881}     & \textbf{0.864} & \textbf{0.883}   & \textbf{0.879}     \\
  \hline
              & M-detector w/ front-end only Odom       & {0.227}     & {0.109} & {0.243}   & {0.177}     \\
              & Removert w/ front-end only Odom         & 0.020     & 0.017 & 0.180   & 0.033     \\
  Sequence 4  & M-detector w/ FAST-LIO                  & {0.664}     & \underline{0.604} & \underline{0.814}   & \underline{0.697}     \\
              & Removert w/ A-LOAM                      & \underline{0.684}     & 0.408 & 0.525   & 0.562     \\
              & Ours w/ front-end only Odom             & \textbf{0.911}     & \textbf{0.853} & \textbf{0.883}   & \textbf{0.887}     \\
  \hline
  \end{tblr}
  \label{tab:metrics}
\end{table*}

\subsection{Experimental Results}
We evaluated the performance of the proposed method using the collected dataset
and compared it with two state-of-the-art methods: M-detector
\cite{wu2024moving} with FAST-LIO \cite{xu2022fast} and Removert
\cite{kim2020remove} with a modified A-LOAM \cite{kim2023scaloam}. FAST-LIO and
A-LOAM provide the robot's movement information to M-detector and Removert,
respectively. Additionally, we evaluated the performance of M-detector and
Removert using the same front-end-only odometry as our method. The
front-end-only odometry drifts significantly compared to FAST-LIO and A-LOAM.

\myold{The results are presented in Table \ref{tab:metrics}.}
\mynew{From the data presented in Table \ref{tab:metrics}, our method with
front-end-only odometry consistently achieves the best performance across all
sequences and metrics. The second-best performance is often achieved by the
M-detector with FAST-LIO, which shows strong results, particularly in Precision
and Recall. Removert with front-end-only odometry consistently
underperforms across all metrics and sequences.}
\myold{In Sequence 1}\mynew{Taking Sequence 1 in details}, M-detector with front-end-only odometry shows a Precision of
0.471 and an IoU of 0.455, with a relatively high Recall of 0.837, but the
overall performance is mediocre \mynew{due to the drift in odometry, which 
negatively impacts map quality and the ability to examine occlusions.}\todo{say why}. 
Removert with front-end-only
odometry performs poorly across all metrics. M-detector with FAST-LIO
achieves a lower Precision of 0.695 and a very high Recall of 0.915, with the
second-best IoU of 0.669, indicating good performance, but with many false
positive points. Removert with A-LOAM has moderate performance with
balanced metrics, and its high precision with low recall and IoU means it can
detect dynamic objects accurately, but it also misses many dynamic objects. The
best performance is from our method with front-end-only odometry, which excels
across all metrics with a Precision of 0.871, an IoU of 0.842, a Recall of
0.898, and an F1 Score of 0.875.

\todo[inline]{I don't think you need to commend on all the sequences unless
  there is something new happening. Better to give aggregate results.}
\myold{
In Sequence 2, the M-detector with front-end only Odom method has a Precision of
0.486, a moderate IoU of 0.452, and a Recall of 0.751. The Removert with
front-end only Odom method again shows very poor performance in all metrics. The
M-detector with FAST-LIO method achieves a high Precision of 0.775, an IoU of
0.677, and a Recall of 0.800. The Removert with A-LOAM method demonstrates good
balance with moderate scores in all metrics. The best overall performance is
from our method with front-end only Odom, which has high Precision of 0.883, an
IoU of 0.876, a Recall of 0.899, and an F1 Score of 0.888.

In Sequence 3, the M-detector with front-end only Odom method shows moderate performance with better Recall (0.871) than other metrics. The Removert with front-end only Odom method continues to have low scores across all metrics. The M-detector with FAST-LIO method achieves high Precision (0.801), an IoU of 0.728, and a Recall of 0.844, with the second-best F1 Score of 0.808. The Removert with A-LOAM method has moderate Precision (0.787) but lower Recall (0.421). Our method with front-end only Odom once again delivers the best results across all metrics with a Precision of 0.881, an IoU of 0.864, a Recall of 0.883, and an F1 Score of 0.879.

In Sequence 4, the M-detector with front-end only Odom method shows poor performance across all metrics. The Removert with front-end only Odom method has very low scores. The M-detector with FAST-LIO method has moderate performance with high Recall (0.814). The Removert with A-LOAM method demonstrates balanced metrics with moderate scores. The Ours with front-end only Odom method achieves the highest scores in Precision (0.911), an IoU of 0.853, and high F1 Score (0.887).}

\myold{From the data presented in Table \ref{tab:metrics}, our method with
front-end-only odometry consistently achieves the best performance across all
sequences and metrics. The second-best performance is often achieved by the
M-detector with FAST-LIO, which shows strong results, particularly in Precision
and Recall. Removert with front-end-only odometry consistently
underperforms across all metrics and sequences. }

Notably, the proposed method relies solely on front-end odometry. The front-end
odometry experiences significant drift during robot movement. However, the
proposed method still achieves the best performance. This is because the
proposed method accumulates the transformation matrix between the current frame
and the initial frame within a time window, allowing it to ignore historical
drift and account only for the drift within the current time window. Conversely,
the other two methods using front-end only odometry perform the worst in all
metrics. This is because front-end-only odometry leads to significant drift,
affecting the accuracy of mapping. However, M-detector and Removert rely on
highly accurate maps to determine if a point is blocked or calculate residuals.

In summary, while all methods suffer from the drift caused by front-end-only
odometry, the proposed method's strategy of focusing on a limited time window
allows it to mitigate the impact of this drift effectively. This approach
significantly enhances its ability to accurately label dynamic objects, as
evidenced by its performance across all evaluated metrics and
sequences. The reliance of M-detector and Removert on precise maps further
exacerbates the performance degradation caused by odometry drift, leading to
their lower performance in dynamic detection tasks.

\subsection{Time Cost}
\begin{figure}[h]
  \centering
  \includegraphics[width=0.5\textwidth]{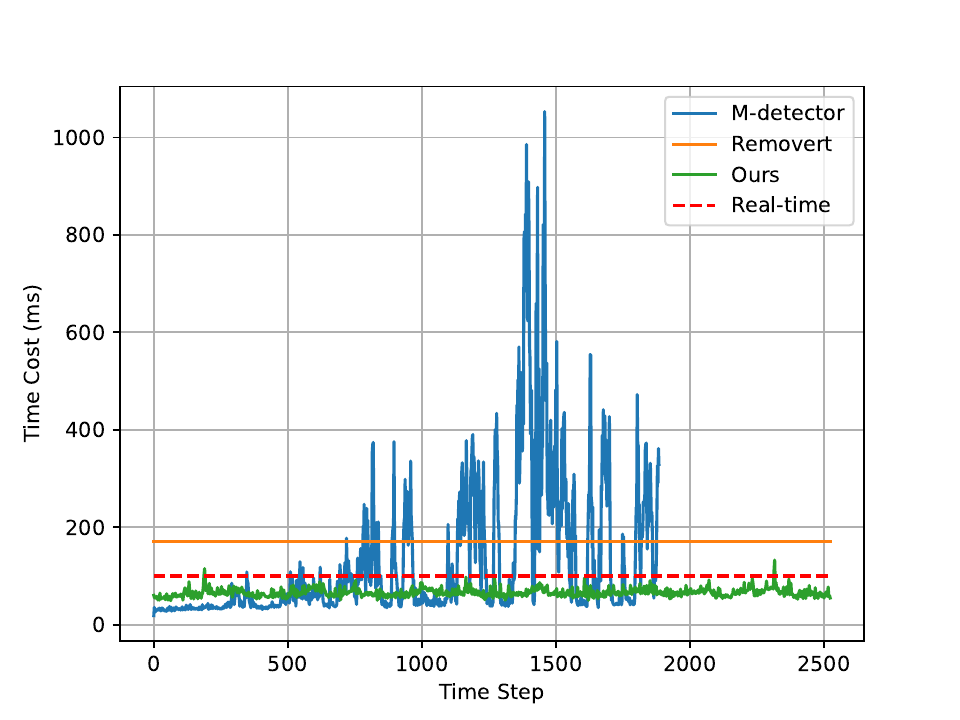}
  \caption{Time Cost of Different Methods: The proposed method integrates front-end odometry and achieves real-time performance. In contrast, the other two methods rely on odometry from external algorithms, which is not included in the time cost of M-detector and Removert. Among the three methods, M-detector with FAST-LIO is the fastest for the first 500 frames, but its long-term performance does not match that of the proposed method.}
  \label{fig:time_cost}
\end{figure}

The time cost comparison of the three methods is shown in Figure
\ref{fig:time_cost}. The proposed method integrates front-end odometry and
achieves real-time performance. In contrast, the other two methods rely on
odometry from external algorithms, and we did not include the external
algorithms' time cost in the time cost of M-detector and Removert. Since
Removert operates offline, it first uses A-LOAM to generate
movement information and scans of the robot's keyframes to files. It then uses
these files to build a map and calculate the residuals between scans and the
map.


The proposed method maintains a consistently low time cost, well below the
real-time threshold throughout the entire time span. This consistent performance
highlights the efficiency of the proposed method in integrating front-end
odometry and achieving real-time processing.

The red dashed line marks the real-time performance threshold. Both M-detector
and Removert frequently exceed this threshold, indicating challenges in
maintaining real-time performance. In contrast, the proposed method remains
consistently below this line, indicating robust real-time performance.

In summary, the proposed method significantly outperforms M-detector and
Removert in terms of time cost, maintaining real-time performance throughout the
observed period. M-detector shows considerable variability and occasional spikes
that disrupt real-time processing, while Removert, despite being more stable
than M-detector, consistently exceeds the real-time threshold. These
observations underscore the efficiency and stability of the proposed method in
dynamic object labeling tasks, leveraging front-end odometry to achieve superior
real-time performance.

\todo[inline]{Discussion is short, you can merge with conclusion --Done}

\section{Conclusion}
\label{sec:conclusion}
In this paper, we propose a real-time dynamic object detection and tracking method for mobile robots equipped with LiDAR sensors. Our approach leverages front-end odometry to enable real-time detection of dynamic objects. By converting unordered LiDAR point cloud data into intensity images and applying a Gaussian Convolution Kernel, our method effectively segments feature points of foreground objects. These segmented points are then clustered into distinct objects and tracked within a sliding time window. This technique accumulates the transformation matrix between the current frame and the initial frame, allowing it to disregard historical drift and focus solely on the drift within the current window.

Our experiments demonstrate that this method achieves superior performance in dynamic object detection across four test sequences, significantly outperforming other state-of-the-art methods. The proposed system shows strong resilience to odometry drift, maintaining high detection accuracy and recall rates. Furthermore, it achieves real-time processing, ensuring reliable operation in dynamic environments over extended periods.

This solution presents a robust and efficient approach to real-time dynamic object detection and tracking, enabling mobile robots to navigate safely and effectively in environments with moving objects. By enhancing the system's capability to manage both static and dynamic features, our method contributes to the development of more accurate and reliable SLAM systems for autonomous robotics.

However, the proposed method performs well in normal environments but is sensitive to
object occlusion. When objects are blocked by other objects, the method loses track
of them in the tracking process. Upon reappearance, the method treats them
as new objects and tracks them from the initial frame. To address this issue, we
can implement a data association method to compare the objects in the current
frame with those in the previous 2 or 3 frames. If matches are found, we can
maintain continuous tracking of the object from the initial frame.
Alternatively, a Kalman filter \cite{kalman1960new} can be used to predict the
object's position when it is occluded. This approach will improve tracking
accuracy and reduce tracking loss.

In addition to data association and Kalman filtering, exploring advanced feature matching techniques, such as deep learning-based object re-identification, could further enhance the robustness of the method against occlusion. By combining these strategies, the proposed method can achieve more reliable and precise tracking in dynamic environments, ultimately contributing to the development of more effective SLAM systems for various robotic applications.

Furthermore, integrating the proposed method into a SLAM (Simultaneous Localization and Mapping) system can enhance localization accuracy. By filtering out dynamic objects in the back-end of the SLAM system, we can eliminate ghost map points and refine the generated map. This integration will leverage the SLAM system's ability to manage static and dynamic features, leading to more robust and accurate environmental mapping and object tracking.

Overall, this section addresses the limitations of our study and proposes future research directions. We suggest advancements in dynamic object tracking with a unstable ego-motion estimation, discuss the broader implications of our findings, and highlight potential applications of our method. This comprehensive overview underscores our contributions to the field of robotics, particularly in autonomous navigation.




\section*{Acknowledgment}
The authors would like to extend their sincere gratitude to Dong Wang, Konno Genta, and Timothy Lee for their invaluable assistance with the dataset collection. Their expertise and dedication were instrumental in acquiring high-quality data, which was critical to the success of this research. We appreciate their willingness to invest considerable time and effort into this project.

\bibliographystyle{IEEEtran}
\bibliography{IEEEabrv,root}

\begin{thebibliography}{10}
\providecommand{\url}[1]{#1}
\csname url@samestyle\endcsname
\providecommand{\newblock}{\relax}
\providecommand{\bibinfo}[2]{#2}
\providecommand{\BIBentrySTDinterwordspacing}{\spaceskip=0pt\relax}
\providecommand{\BIBentryALTinterwordstretchfactor}{4}
\providecommand{\BIBentryALTinterwordspacing}{\spaceskip=\fontdimen2\font plus
\BIBentryALTinterwordstretchfactor\fontdimen3\font minus \fontdimen4\font\relax}
\providecommand{\BIBforeignlanguage}[2]{{%
\expandafter\ifx\csname l@#1\endcsname\relax
\typeout{** WARNING: IEEEtran.bst: No hyphenation pattern has been}%
\typeout{** loaded for the language `#1'. Using the pattern for}%
\typeout{** the default language instead.}%
\else
\language=\csname l@#1\endcsname
\fi
#2}}
\providecommand{\BIBdecl}{\relax}
\BIBdecl

\bibitem{campos2021orb}
C.~Campos, R.~Elvira, J.~J.~G. Rodr{\'\i}guez, J.~M. Montiel, and J.~D. Tard{\'o}s, ``{ORB-SLAM3: An Accurate Open-Source Library for Visual, Visual-Inertial, and Multimap SLAM},'' \emph{IEEE Transactions on Robotics}, 2021.

\bibitem{zhang2014loam}
J.~Zhang and S.~Singh, ``{LOAM}: Lidar odometry and mapping in real-time.'' in \emph{Robotics: Science and systems}, vol.~2, no.~9.\hskip 1em plus 0.5em minus 0.4em\relax Berkeley, CA, 2014, pp. 1--9.

\bibitem{qin2018vins}
T.~Qin, P.~Li, and S.~Shen, ``{VINS-Mono}: A robust and versatile monocular visual-inertial state estimator,'' \emph{IEEE Transactions on Robotics}, vol.~34, no.~4, pp. 1004--1020, 2018.

\bibitem{xu2022fast}
W.~Xu, Y.~Cai, D.~He, J.~Lin, and F.~Zhang, ``{FAST-LIO2}: Fast direct lidar-inertial odometry,'' \emph{IEEE Transactions on Robotics}, 2022.

\bibitem{shan2021lvi}
T.~Shan, B.~Englot, C.~Ratti, and D.~Rus, ``{LVI-SAM: Tightly-coupled Lidar-Visual-Inertial Odometry via Smoothing and Mapping},'' in \emph{2021 IEEE international conference on robotics and automation (ICRA)}.\hskip 1em plus 0.5em minus 0.4em\relax IEEE, 2021, pp. 5692--5698.

\bibitem{liosam2020shan}
T.~Shan, B.~Englot, D.~Meyers, W.~Wang, C.~Ratti, and R.~Daniela, ``{{LIO-SAM}: Tightly-coupled Lidar Inertial Odometry via Smoothing and Mapping},'' in \emph{IEEE/RSJ International Conference on Intelligent Robots and Systems (IROS)}.\hskip 1em plus 0.5em minus 0.4em\relax IEEE, 2020, pp. 5135--5142.

\bibitem{furgale2013toward}
P.~Furgale, U.~Schwesinger, M.~Rufli, W.~Derendarz, H.~Grimmett, P.~M{\"u}hlfellner, S.~Wonneberger, J.~Timpner, S.~Rottmann, B.~Li \emph{et~al.}, ``Toward automated driving in cities using close-to-market sensors: An overview of the v-charge project,'' in \emph{2013 IEEE Intelligent Vehicles Symposium (IV)}.\hskip 1em plus 0.5em minus 0.4em\relax IEEE, 2013, pp. 809--816.

\bibitem{peng2024review}
H.~Peng, Z.~Zhao, and L.~Wang, ``{A Review of Dynamic Object Filtering in SLAM Based on 3D LiDAR},'' \emph{Sensors}, vol.~24, no.~2, p. 645, 2024.

\bibitem{redmon2016you}
J.~Redmon, S.~Divvala, R.~Girshick, and A.~Farhadi, ``{You Only Look Once: Unified, Real-Time Object Detection},'' in \emph{Proceedings of the IEEE conference on computer vision and pattern recognition}, 2016, pp. 779--788.

\bibitem{redmon2017yolo9000}
J.~Redmon and A.~Farhadi, ``{YOLO9000: better, faster, stronger},'' in \emph{Proceedings of the IEEE conference on computer vision and pattern recognition}, 2017, pp. 7263--7271.

\bibitem{redmon2018yolov3}
A.~Farhadi and J.~Redmon, ``{YOLOv3: An incremental improvement},'' in \emph{Computer vision and pattern recognition}, vol. 1804.\hskip 1em plus 0.5em minus 0.4em\relax Springer Berlin/Heidelberg, Germany, 2018, pp. 1--6.

\bibitem{bochkovskiy2020yolov4}
A.~Bochkovskiy, C.-Y. Wang, and H.-Y.~M. Liao, ``{YOLOv4: Optimal speed and accuracy of object detection},'' \emph{arXiv preprint arXiv:2004.10934}, 2020.

\bibitem{ge2021yolox}
Z.~Ge, S.~Liu, F.~Wang, Z.~Li, and J.~Sun, ``{YOLOX: Exceeding YOLO Series in 2021},'' \emph{arXiv preprint arXiv:2107.08430}, 2021.

\bibitem{lim2021erasor}
H.~Lim, S.~Hwang, and H.~Myung, ``Erasor: Egocentric ratio of pseudo occupancy-based dynamic object removal for static 3d point cloud map building,'' \emph{IEEE Robotics and Automation Letters}, vol.~6, no.~2, pp. 2272--2279, 2021.

\bibitem{arani2022comprehensive}
E.~Arani, S.~Gowda, R.~Mukherjee, O.~Magdy, S.~Kathiresan, and B.~Zonooz, ``A comprehensive study of real-time object detection networks across multiple domains: A survey,'' \emph{arXiv preprint arXiv:2208.10895}, 2022.

\bibitem{schmid2023dynablox}
L.~Schmid, O.~Andersson, A.~Sulser, P.~Pfreundschuh, and R.~Siegwart, ``Dynablox: Real-time detection of diverse dynamic objects in complex environments,'' \emph{IEEE Robotics and Automation Letters}, 2023.

\bibitem{du2023real}
W.~Du and G.~Beltrame, ``Real-time simultaneous localization and mapping with lidar intensity,'' in \emph{2023 IEEE International Conference on Robotics and Automation (ICRA)}.\hskip 1em plus 0.5em minus 0.4em\relax IEEE, 2023, pp. 4164--4170.

\bibitem{wu2024moving}
H.~Wu, Y.~Li, W.~Xu, F.~Kong, and F.~Zhang, ``Moving event detection from lidar point streams,'' \emph{Nature Communications}, vol.~15, no.~1, p. 345, 2024.

\bibitem{kim2020remove}
G.~Kim and A.~Kim, ``Remove, then revert: Static point cloud map construction using multiresolution range images,'' in \emph{2020 IEEE/RSJ International Conference on Intelligent Robots and Systems (IROS)}.\hskip 1em plus 0.5em minus 0.4em\relax IEEE, 2020, pp. 10\,758--10\,765.

\bibitem{azim2012detection}
A.~Azim and O.~Aycard, ``Detection, classification and tracking of moving objects in a 3d environment,'' in \emph{2012 IEEE Intelligent Vehicles Symposium}.\hskip 1em plus 0.5em minus 0.4em\relax IEEE, 2012, pp. 802--807.

\bibitem{schauer2018peopleremover}
J.~Schauer and A.~N{\"u}chter, ``The peopleremover—removing dynamic objects from 3-d point cloud data by traversing a voxel occupancy grid,'' \emph{IEEE robotics and automation letters}, vol.~3, no.~3, pp. 1679--1686, 2018.

\bibitem{thrun2003learning}
S.~Thrun, ``Learning occupancy grid maps with forward sensor models,'' \emph{Autonomous robots}, vol.~15, pp. 111--127, 2003.

\bibitem{hornung2013octomap}
A.~Hornung, K.~M. Wurm, M.~Bennewitz, C.~Stachniss, and W.~Burgard, ``Octomap: An efficient probabilistic 3d mapping framework based on octrees,'' \emph{Autonomous robots}, vol.~34, pp. 189--206, 2013.

\bibitem{schmid2024khronos}
L.~Schmid, M.~Abate, Y.~Chang, and L.~Carlone, ``Khronos: A unified approach for spatio-temporal metric-semantic slam in dynamic environments,'' \emph{arXiv preprint arXiv:2402.13817}, 2024.

\bibitem{lim2023erasor2}
H.~Lim, L.~Nunes, B.~Mersch, X.~Chen, J.~Behley, H.~Myung, and C.~Stachniss, ``Erasor2: Instance-aware robust 3d mapping of the static world in dynamic scenes,'' in \emph{Robotics: Science and Systems (RSS 2023)}.\hskip 1em plus 0.5em minus 0.4em\relax IEEE, 2023.

\bibitem{pomerleau2014long}
F.~Pomerleau, P.~Kr{\"u}si, F.~Colas, P.~Furgale, and R.~Siegwart, ``Long-term 3d map maintenance in dynamic environments,'' in \emph{2014 IEEE International Conference on Robotics and Automation (ICRA)}.\hskip 1em plus 0.5em minus 0.4em\relax IEEE, 2014, pp. 3712--3719.

\bibitem{banerjee2019lifelong}
N.~Banerjee, D.~Lisin, J.~Briggs, M.~Llofriu, and M.~E. Munich, ``Lifelong mapping using adaptive local maps,'' in \emph{2019 European Conference on Mobile Robots (ECMR)}.\hskip 1em plus 0.5em minus 0.4em\relax IEEE, 2019, pp. 1--8.

\bibitem{ambrucs2014meta}
R.~Ambru{\c{s}}, N.~Bore, J.~Folkesson, and P.~Jensfelt, ``Meta-rooms: Building and maintaining long term spatial models in a dynamic world,'' in \emph{2014 IEEE/RSJ international conference on intelligent robots and systems}.\hskip 1em plus 0.5em minus 0.4em\relax IEEE, 2014, pp. 1854--1861.

\bibitem{jiang2016static}
C.~Jiang, D.~P. Paudel, Y.~Fougerolle, D.~Fofi, and C.~Demonceaux, ``Static-map and dynamic object reconstruction in outdoor scenes using 3-d motion segmentation,'' \emph{IEEE Robotics and Automation Letters}, vol.~1, no.~1, pp. 324--331, 2016.

\bibitem{palazzolo2019refusion}
E.~Palazzolo, J.~Behley, P.~Lottes, P.~Giguere, and C.~Stachniss, ``Refusion: 3d reconstruction in dynamic environments for rgb-d cameras exploiting residuals,'' in \emph{2019 IEEE/RSJ International Conference on Intelligent Robots and Systems (IROS)}.\hskip 1em plus 0.5em minus 0.4em\relax IEEE, 2019, pp. 7855--7862.

\bibitem{curless1996volumetric}
B.~Curless and M.~Levoy, ``A volumetric method for building complex models from range images,'' in \emph{Proceedings of the 23rd annual conference on Computer graphics and interactive techniques}, 1996, pp. 303--312.

\bibitem{nunes2022unsupervised}
L.~Nunes, X.~Chen, R.~Marcuzzi, A.~Osep, L.~Leal-Taix{\'e}, C.~Stachniss, and J.~Behley, ``Unsupervised class-agnostic instance segmentation of 3d lidar data for autonomous vehicles,'' \emph{IEEE Robotics and Automation Letters}, vol.~7, no.~4, pp. 8713--8720, 2022.

\bibitem{thomas2022learning}
H.~Thomas, M.~G. de~Saint~Aurin, J.~Zhang, and T.~D. Barfoot, ``Learning spatiotemporal occupancy grid maps for lifelong navigation in dynamic scenes,'' in \emph{2022 International Conference on Robotics and Automation (ICRA)}.\hskip 1em plus 0.5em minus 0.4em\relax IEEE, 2022, pp. 484--490.

\bibitem{thomas2019kpconv}
H.~Thomas, C.~R. Qi, J.-E. Deschaud, B.~Marcotegui, F.~Goulette, and L.~J. Guibas, ``Kpconv: Flexible and deformable convolution for point clouds,'' in \emph{Proceedings of the IEEE/CVF international conference on computer vision}, 2019, pp. 6411--6420.

\bibitem{ronneberger2015u}
O.~Ronneberger, P.~Fischer, and T.~Brox, ``U-net: Convolutional networks for biomedical image segmentation,'' in \emph{Medical image computing and computer-assisted intervention--MICCAI 2015: 18th international conference, Munich, Germany, October 5-9, 2015, proceedings, part III 18}.\hskip 1em plus 0.5em minus 0.4em\relax Springer, 2015, pp. 234--241.

\bibitem{chen2021moving}
X.~Chen, S.~Li, B.~Mersch, L.~Wiesmann, J.~Gall, J.~Behley, and C.~Stachniss, ``Moving object segmentation in 3d lidar data: A learning-based approach exploiting sequential data,'' \emph{IEEE Robotics and Automation Letters}, vol.~6, no.~4, pp. 6529--6536, 2021.

\bibitem{milioto2019rangenet++}
A.~Milioto, I.~Vizzo, J.~Behley, and C.~Stachniss, ``Rangenet++: Fast and accurate lidar semantic segmentation,'' in \emph{2019 IEEE/RSJ international conference on intelligent robots and systems (IROS)}.\hskip 1em plus 0.5em minus 0.4em\relax IEEE, 2019, pp. 4213--4220.

\bibitem{li2021multi}
S.~Li, X.~Chen, Y.~Liu, D.~Dai, C.~Stachniss, and J.~Gall, ``Multi-scale interaction for real-time lidar data segmentation on an embedded platform,'' \emph{IEEE Robotics and Automation Letters}, vol.~7, no.~2, pp. 738--745, 2021.

\bibitem{cortinhal2020salsanext}
T.~Cortinhal, G.~Tzelepis, and E.~Erdal~Aksoy, ``Salsanext: Fast, uncertainty-aware semantic segmentation of lidar point clouds,'' in \emph{Advances in Visual Computing: 15th International Symposium, ISVC 2020, San Diego, CA, USA, October 5--7, 2020, Proceedings, Part II 15}.\hskip 1em plus 0.5em minus 0.4em\relax Springer, 2020, pp. 207--222.

\bibitem{sun2022efficient}
J.~Sun, Y.~Dai, X.~Zhang, J.~Xu, R.~Ai, W.~Gu, and X.~Chen, ``Efficient spatial-temporal information fusion for lidar-based 3d moving object segmentation,'' in \emph{2022 IEEE/RSJ International Conference on Intelligent Robots and Systems (IROS)}.\hskip 1em plus 0.5em minus 0.4em\relax IEEE, 2022, pp. 11\,456--11\,463.

\bibitem{mersch2022receding}
B.~Mersch, X.~Chen, I.~Vizzo, L.~Nunes, J.~Behley, and C.~Stachniss, ``Receding moving object segmentation in 3d lidar data using sparse 4d convolutions,'' \emph{IEEE Robotics and Automation Letters}, vol.~7, no.~3, pp. 7503--7510, 2022.

\bibitem{choy20194d}
C.~Choy, J.~Gwak, and S.~Savarese, ``4d spatio-temporal convnets: Minkowski convolutional neural networks,'' in \emph{Proceedings of the IEEE/CVF conference on computer vision and pattern recognition}, 2019, pp. 3075--3084.

\bibitem{rublee2011orb}
E.~Rublee, V.~Rabaud, K.~Konolige, and G.~Bradski, ``Orb: An efficient alternative to sift or surf,'' in \emph{2011 International conference on computer vision}.\hskip 1em plus 0.5em minus 0.4em\relax Ieee, 2011, pp. 2564--2571.

\bibitem{bar1995multitarget}
Y.~Bar-Shalom and X.-R. Li, \emph{Multitarget-multisensor tracking: principles and techniques}.\hskip 1em plus 0.5em minus 0.4em\relax YBS publishing Storrs, CT, 1995, vol.~19.

\bibitem{kuhn1955hungarian}
H.~W. Kuhn, ``The hungarian method for the assignment problem,'' \emph{Naval research logistics quarterly}, vol.~2, no. 1-2, pp. 83--97, 1955.

\bibitem{goutte2005probabilistic}
C.~Goutte and E.~Gaussier, ``A probabilistic interpretation of precision, recall and f-score, with implication for evaluation,'' in \emph{European conference on information retrieval}.\hskip 1em plus 0.5em minus 0.4em\relax Springer, 2005, pp. 345--359.

\bibitem{kim2023scaloam}
G.~Kim, ``{SC-A-LOAM},'' \url{https://github.com/gisbi-kim/SC-A-LOAM}, 2021, accessed: 2021-07-16.

\bibitem{kalman1960new}
R.~E. Kalman, ``A new approach to linear filtering and prediction problems,'' 1960.

\bibitem{wilcoxon1992individual}
F.~Wilcoxon, ``Individual comparisons by ranking methods,'' in \emph{Breakthroughs in statistics: Methodology and distribution}.\hskip 1em plus 0.5em minus 0.4em\relax Springer, 1992, pp. 196--202.

\end{thebibliography}

\newpage
\clearpage

\appendix
\renewcommand\thefigure{\thesection.\arabic{figure}}
\setcounter{figure}{0} 

\section{Appendix Section}

\subsection{Detailed Analysis of Experimental Results}

\begin{figure*}[ht!]
  \centering
  \captionsetup[subfloat]{font=small, labelfont=bf}
  \subfloat[\textit{Precision results in Sequence 1}: This subfigure shows the precision results for the first sequence. The precision of each method is plotted, demonstrating the variability and performance across the sequence. Our method with front-end odometry achieves the highest and most stable precision throughout the sequence, significantly outperforming the other methods. M-detector with FAST-LIO also shows good performance but with more variability. Removert and M-detector with front-end odometry show lower precision.
  ]{%
    \includegraphics[width=0.48\textwidth]{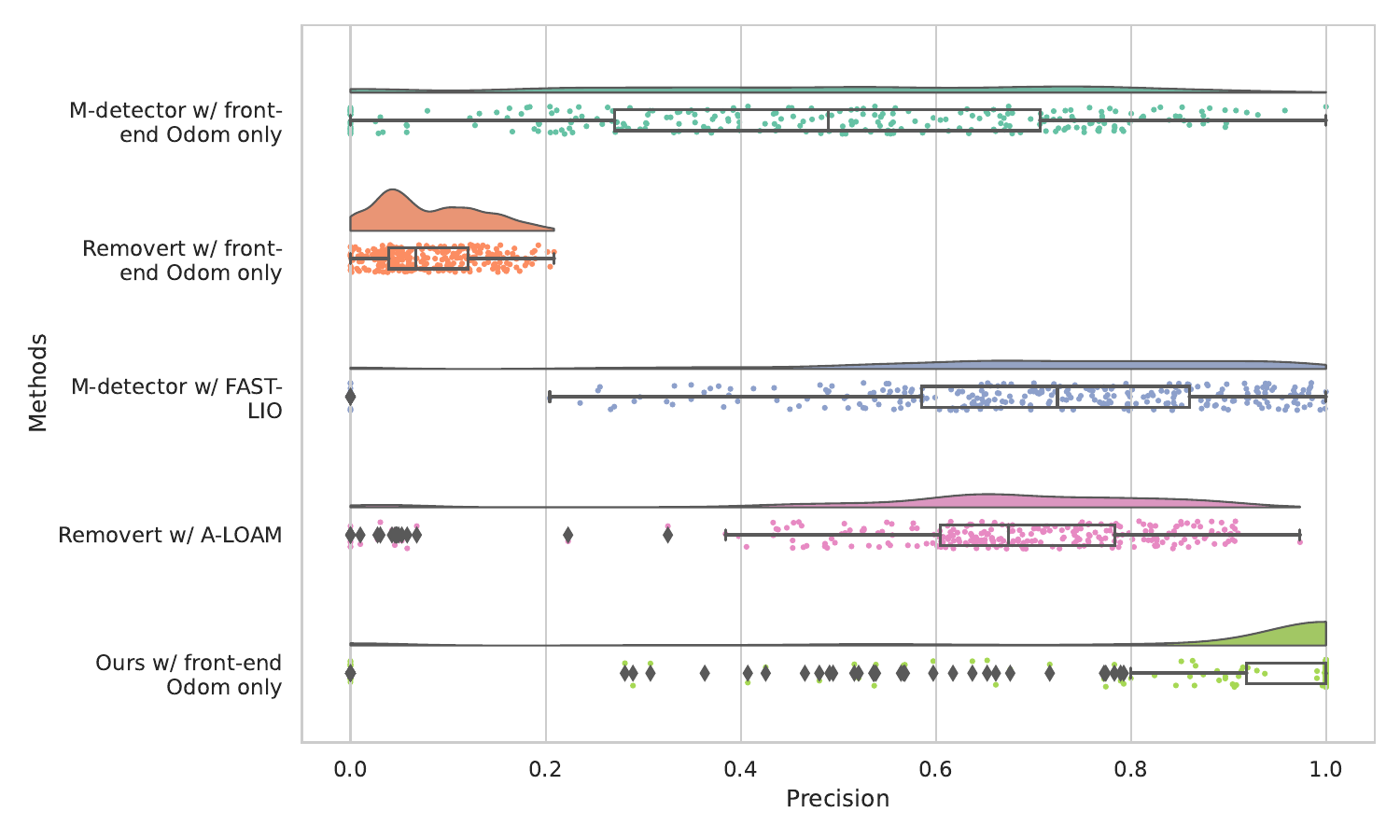}%
    \label{fig:precision1}%
  }\hfill
  \subfloat[\textit{Precision results in Sequence 2}: This subfigure displays the precision results for the second sequence. Similar to Sequence 1, our method with front-end odometry maintains the highest precision with minimal variability. M-detector with FAST-LIO again shows high precision but with notable fluctuations. Removert and M-detector with front-end odometry continue to perform poorly, indicating a consistent trend across different sequences.]{%
    \includegraphics[width=0.48\textwidth]{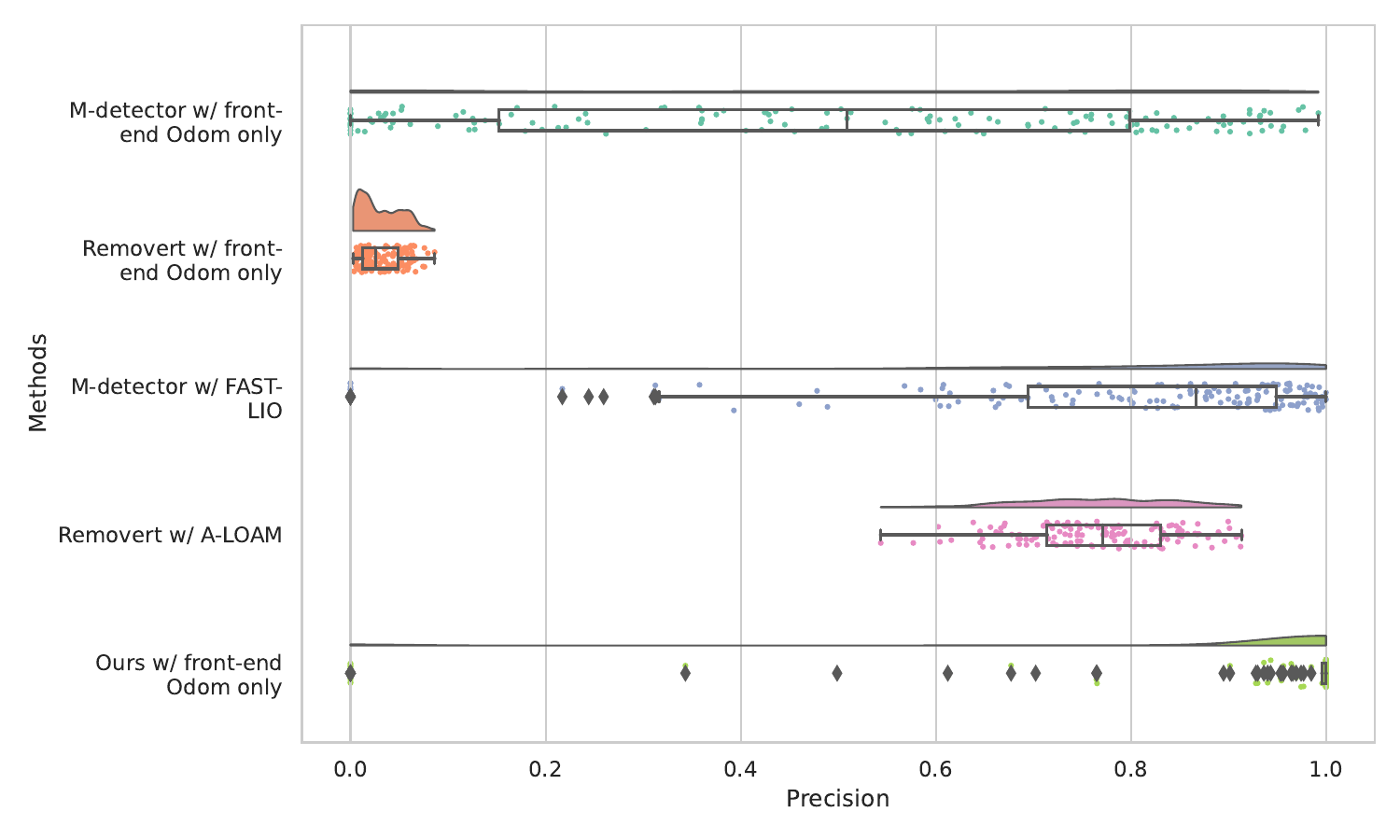}%
    \label{fig:precision2}%
  }

  \vspace{0.1cm}

  \subfloat[\textit{Precision results in Sequence 3}: This subfigure presents the precision results for the third sequence. It highlights how each method performed in terms of precision throughout the sequence. Our method consistently achieves the highest precision, reinforcing its robustness and reliability. The performance of M-detector with FAST-LIO remains high but less stable compared to our method. Removert with A-LOAM has slightly lower precision to M-detector with FAST-LIO, but has more stable performance. Removert with front-end odometry exhibit similar lower precision trends as observed in previous sequences.]{%
    \includegraphics[width=0.48\textwidth]{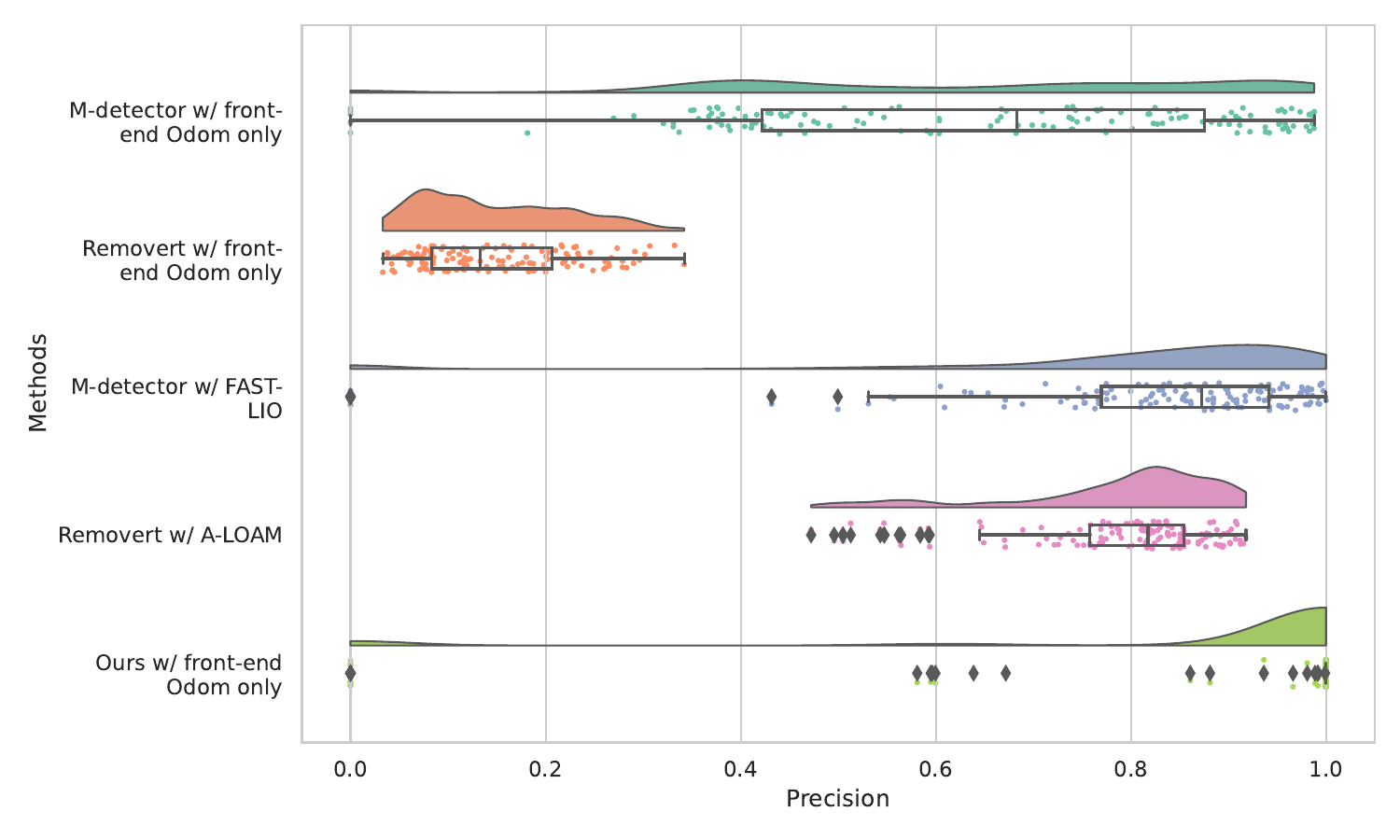}%
    \label{fig:precision3}%
  }\hfill
  \subfloat[\textit{Precision results in Sequence 4}: This subfigure illustrates the precision results for the fourth sequence. It offers insights into the precision performance of each method in this sequence. Our method with front-end odometry once again leads with the highest precision and lowest variability. M-detector with FAST-LIO shows competitive precision but with occasional dips. Removert with A-LOAM still has similar performance with M-detector with FAST-LIO, but has more stable performance. Removert and M-detector with front-end odometry show the least precision, consistent with the patterns seen in the earlier sequences.]{%
    \includegraphics[width=0.48\textwidth]{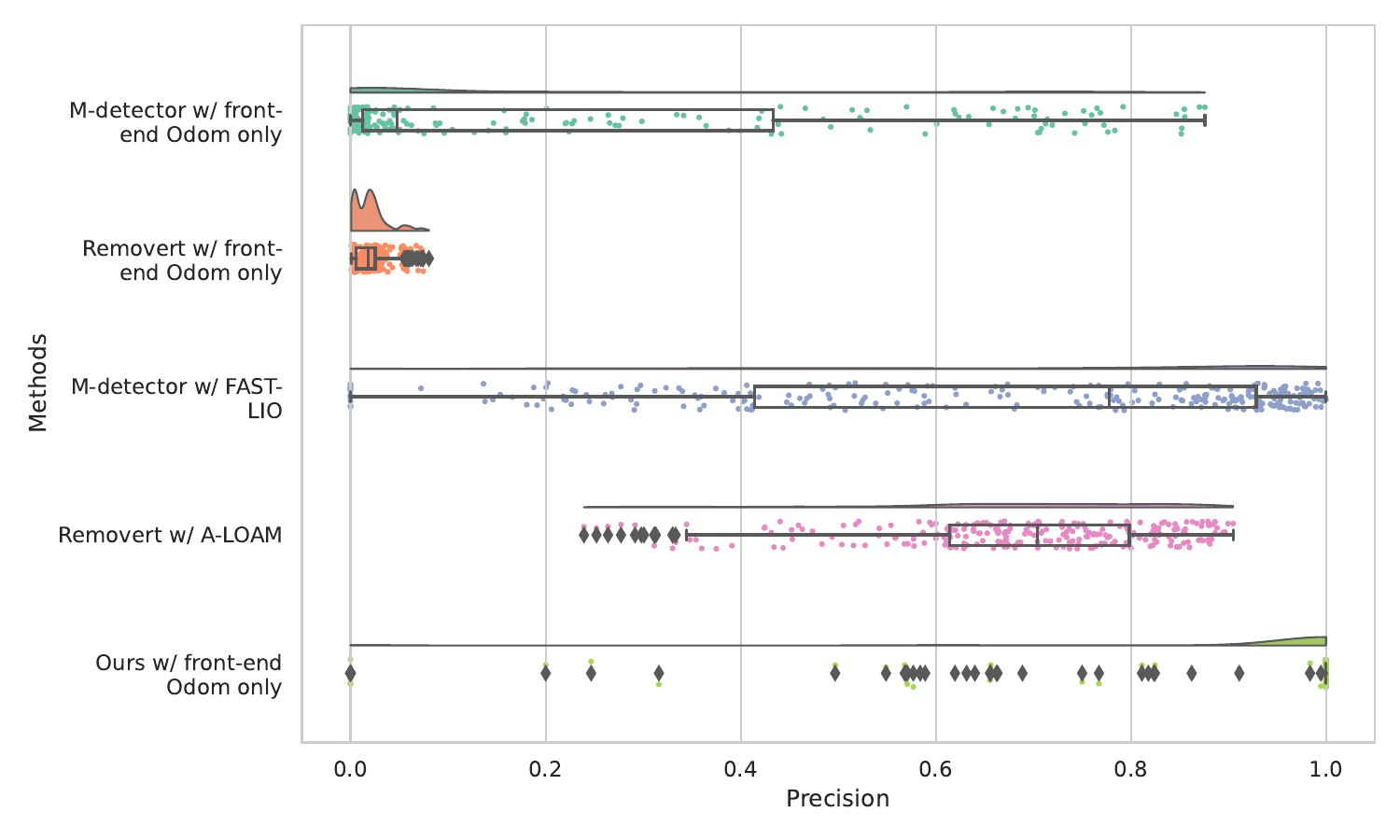}%
    \label{fig:precision4}%
  }
  \caption{\textbf{Precision results across four sequences.} This figure presents the precision results for different dynamic object detection methods across four test sequences. The methods compared are: M-detector with front-end odometry, Removert with front-end odometry, M-detector with FAST-LIO, Removert with A-LOAM, and our method with front-end odometry.}
  \label{fig:precision}
\end{figure*}

The experimental results shown in Figure \ref{fig:precision} provide a detailed comparison of the precision performance of different dynamic object detection methods across four sequences. 
The precision results across four sequences demonstrate the superiority and consistency of our method with front-end odometry. In Sequence 1, our method achieves the highest and most stable precision, significantly outperforming other methods such as M-detector with FAST-LIO and Removert with A-LOAM, which, while performing well, show greater variability. Removert and M-detector with front-end odometry exhibit lower precision and higher variability, indicating less reliability in dynamic object detection.

For Sequence 2, the precision results reinforce the findings from Sequence 1. Our method maintains the highest precision with minimal variability, demonstrating its robustness across different sequences. The performance of M-detector with FAST-LIO, although high in precision, shows notable fluctuations. The performance of Removert with A-LOAM has similar precision with M-detector with FAST-LIO, but lower variability. Meanwhile, Removert with front-end odometry and M-detector with front-end odometry continue to perform poorly, suggesting a consistent trend of lower precision and higher variability.

Sequence 3 precision results highlight the consistency and robustness of our method, which again achieves the highest precision throughout the sequence. M-detector with FAST-LIO remains a strong performer but with less stability compared to our method. The lower precision and higher variability observed in M-detector with front-end odometry, mirror the trends seen in previous sequences, underlining its limitations in dynamic object detection. Removert with A-LOAM in this sequence shows better performance compared to the earlier sequences, but still lags behind our method. Removert with front-end odometry continues to show the least precision, consistent with the earlier sequences, highlighting the challenges in maintaining accuracy with drift odometry.

The precision results in Sequence 4 further validate the effectiveness of our method, which leads with the highest precision and the least variability. M-detector with FAST-LIO shows competitive performance but with occasional dips in precision. Removert and M-detector with front-end odometry continue to show the least precision, consistent with the earlier sequences, highlighting the robustness and reliability of our approach.

\begin{figure*}[ht!]
  \centering
  \captionsetup[subfloat]{font=small, labelfont=bf}
  \subfloat[\textit{IoU results in Sequence 1}: This subfigure presents the IoU results for the first sequence. Our method with front-end odometry achieves the highest IoU, indicating its effectiveness in accurately detecting dynamic objects. M-detector with FAST-LIO shows good IoU but with more variability. Removert with A-LOAM and M-detector with front-end odometry display lower IoU and higher variability, suggesting challenges in maintaining consistency.]{%
    \includegraphics[width=0.48\textwidth]{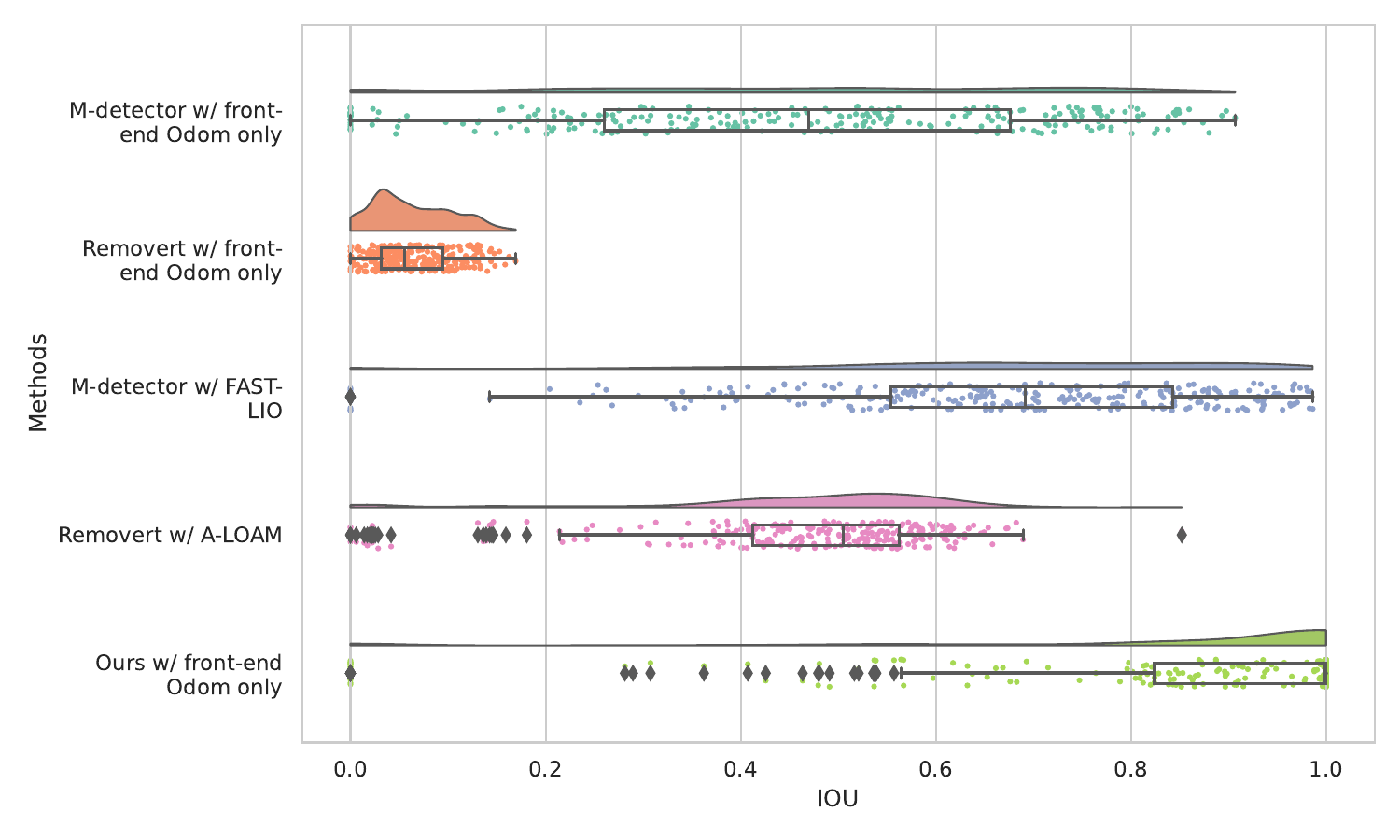}%
    \label{fig:iou1}%
  }\hfill
  \subfloat[\textit{IoU results in Sequence 2}: This subfigure shows the IoU results for the second sequence. Similar to the precision results, our method with front-end odometry achieves the highest IoU with minimal variability. M-detector with FAST-LIO also performs well but shows notable fluctuations. M-detector with front-end odometry continue to show lower IoU and higher variability. Removert has lower variability, but also has lower IoU.]{%
    \includegraphics[width=0.48\textwidth]{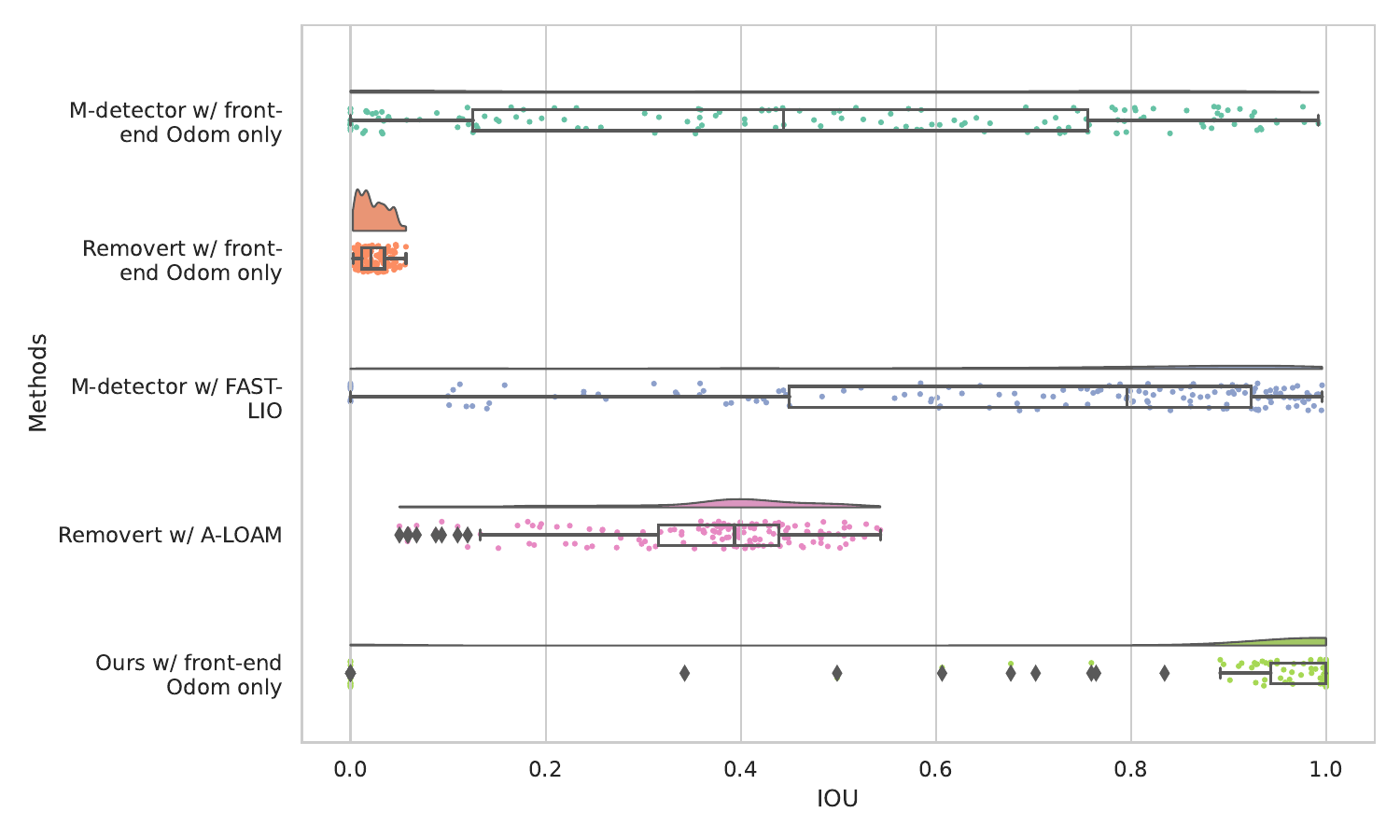}%
    \label{fig:iou2}%
  }

  \vspace{0.1cm} 

  \subfloat[\textit{IoU results in Sequence 3}: This subfigure presents the IoU results for the third sequence. Our method with front-end odometry consistently achieves the highest IoU, reinforcing its robustness. M-detector with FAST-LIO remains competitive but less stable. Removert with A-LOAM and M-detector with front-end odometry show similar lower IoU trends as observed previously.]{%
    \includegraphics[width=0.48\textwidth]{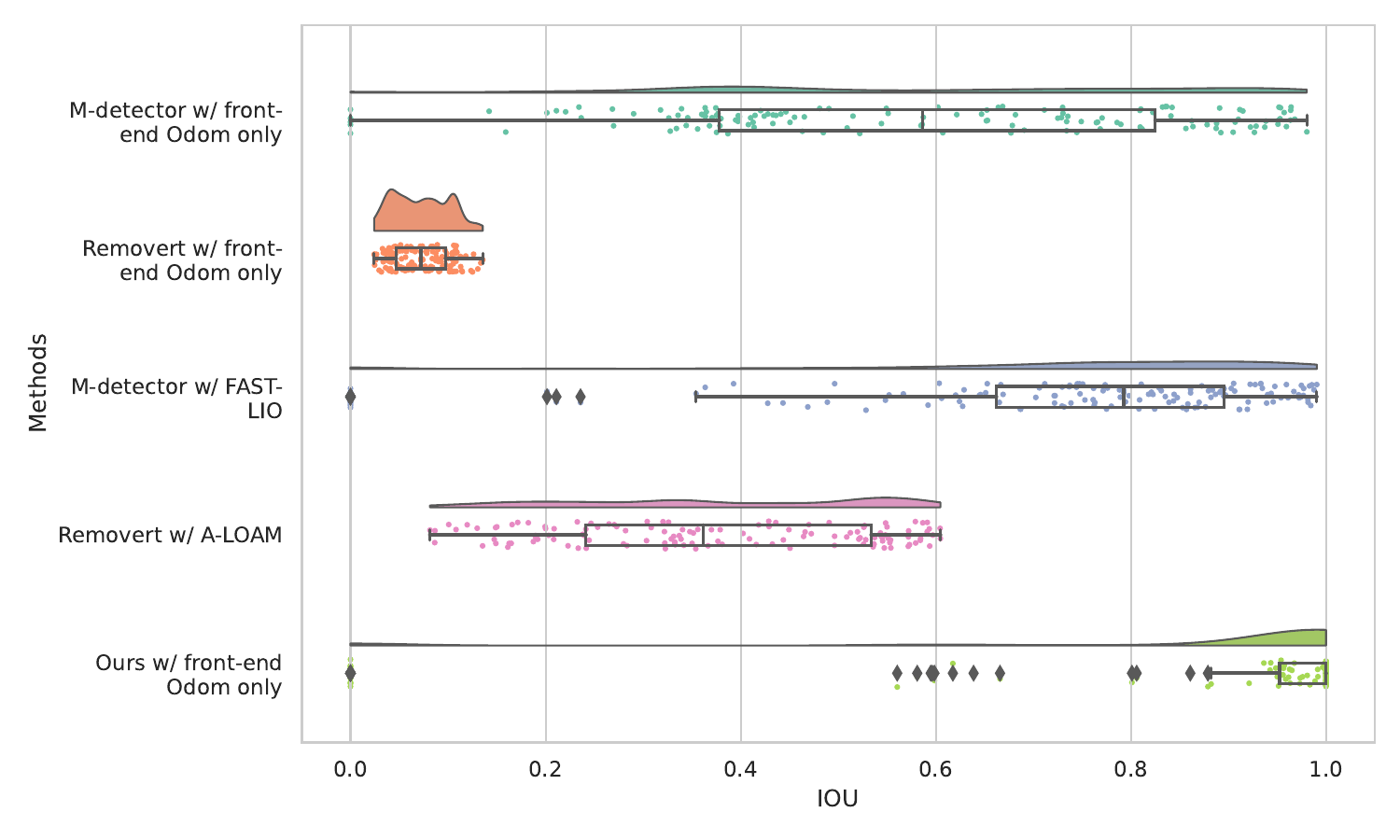}%
    \label{fig:iou3}%
  }\hfill
  \subfloat[\textit{IoU results in Sequence 4}: This subfigure illustrates the IoU results for the fourth sequence. Our method with front-end odometry once again leads with the highest IoU and lowest variability. M-detector with FAST-LIO shows competitive IoU but with occasional dips. Removert and M-detector with front-end odometry show the lowest IoU, consistent with the patterns seen in earlier sequences.]{%
    \includegraphics[width=0.48\textwidth]{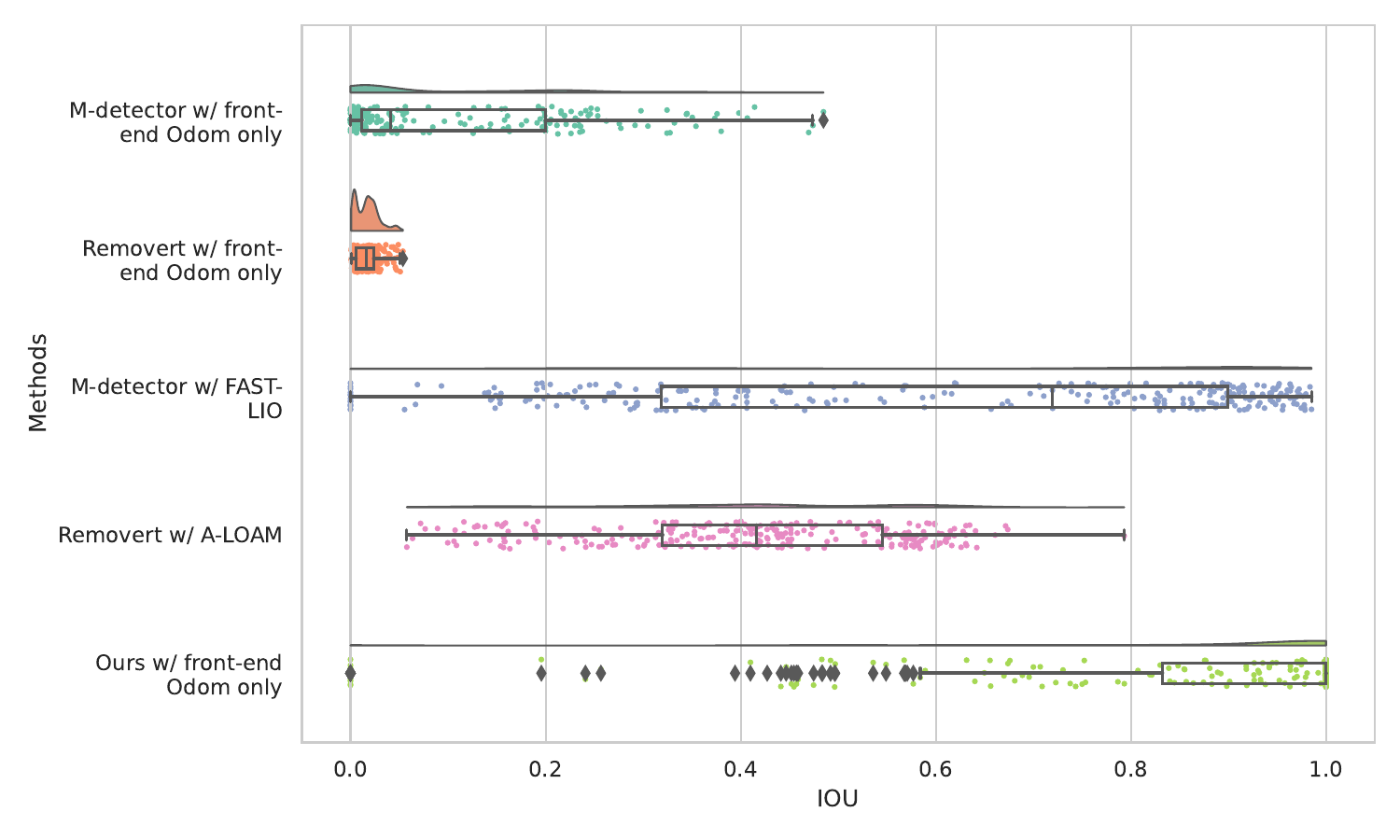}%
    \label{fig:iou4}%
  }
  \caption{\textbf{IoU results across four sequences}. This figure presents the IoU results for different dynamic object detection methods across four test sequences. The methods compared are: M-detector with front-end odometry, Removert with front-end odometry, M-detector with FAST-LIO, Removert with A-LOAM, and our method with front-end odometry.}
  \label{fig:iou}
\end{figure*}

The IoU results in Figure \ref{fig:iou} across the sequences reinforce the trends observed in the precision analysis. In Sequence 1, our method with front-end odometry achieves the highest IoU, demonstrating its effectiveness in accurately detecting dynamic objects. M-detector with FAST-LIO shows good IoU but with greater variability. Removert with A-LOAM and M-detector with front-end odometry display lower IoU and higher variability, indicating challenges in maintaining consistency. The Removert with front-end odometry shows the least IoU among all methods.

In Sequence 2, our method maintains the highest IoU with minimal variability, similar to the precision results. M-detector with FAST-LIO performs well but with notable fluctuations. The lower IoU and higher variability with M-detector with front-end odometry and the least IoU in Removert with front-end odometry, further confirm their limitations in dynamic object detection when they equipped lower precision odometry.

The IoU results for Sequence 3 continue to show our method's robustness, consistently achieving the highest IoU. M-detector with FAST-LIO remains competitive but less stable compared to our method. The trends of lower IoU and higher variability in Removert with A-LOAM and M-detector with front-end odometry, are consistent with previous observations.

For Sequence 4, our method once again leads with the highest IoU and the least variability. M-detector with FAST-LIO shows competitive IoU but with occasional dips. Removert and M-detector with front-end odometry continue to display the lowest IoU, aligning with the patterns seen in earlier sequences.

\begin{figure*}[ht!]
  \centering
  \captionsetup[subfloat]{font=small, labelfont=bf}
  \subfloat[\textit{Recall results in Sequence 1}: This subfigure shows the recall results for the first sequence. M-detector with FAST-LIO achieves the best recall with the least variability. Our method with front-end odometry remains competitive, outperforming Removert methods, which show lower recall and higher variability.]{%
    \includegraphics[width=0.48\textwidth]{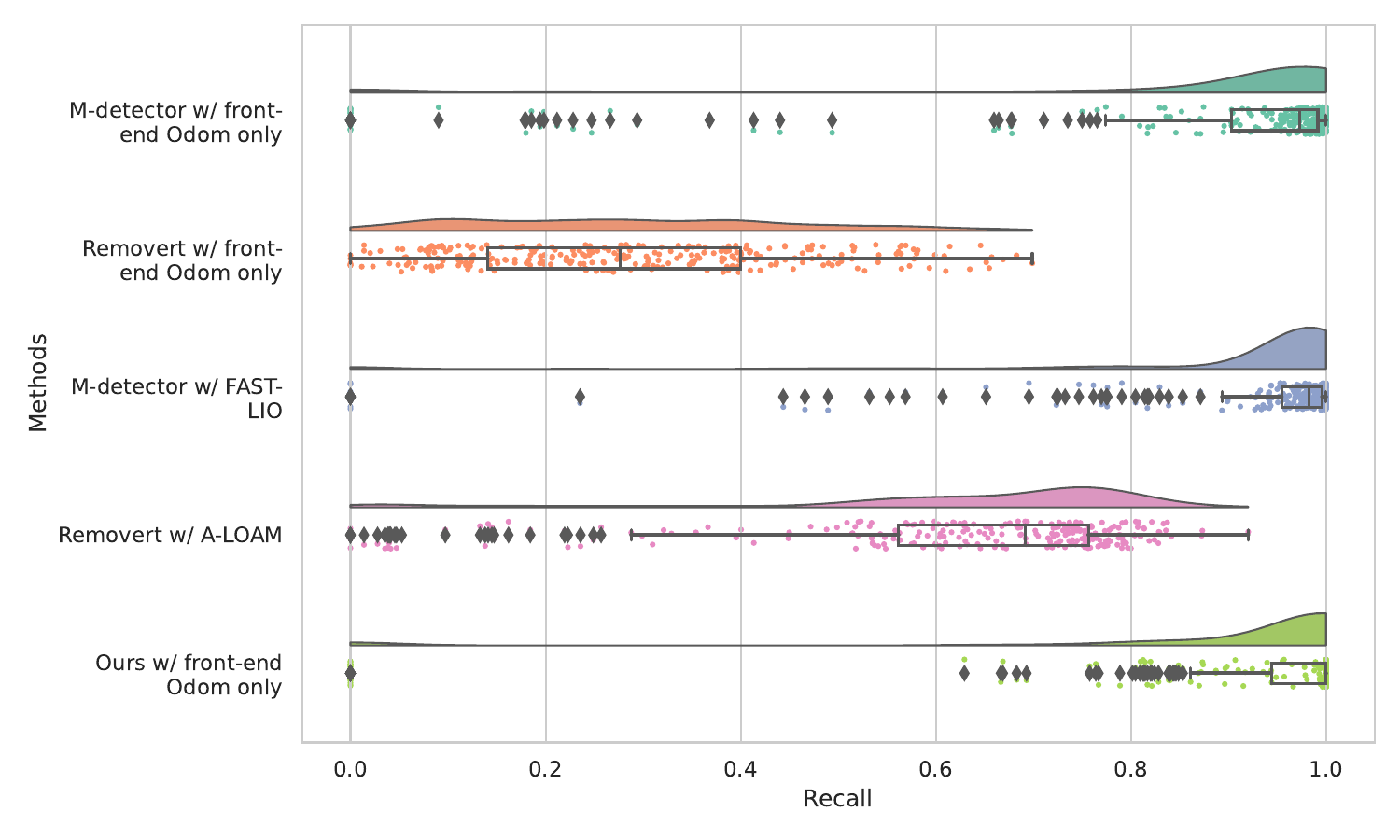}%
    \label{fig:recall1}%
  }\hfill
  \subfloat[\textit{Recall results in Sequence 2}: This subfigure displays the recall results for the second sequence. Our method achieves the highest recall with minimal variability. M-detector with FAST-LIO shows high recall with fluctuations. Removert methods continue to perform poorly with lower recall.]{%
    \includegraphics[width=0.48\textwidth]{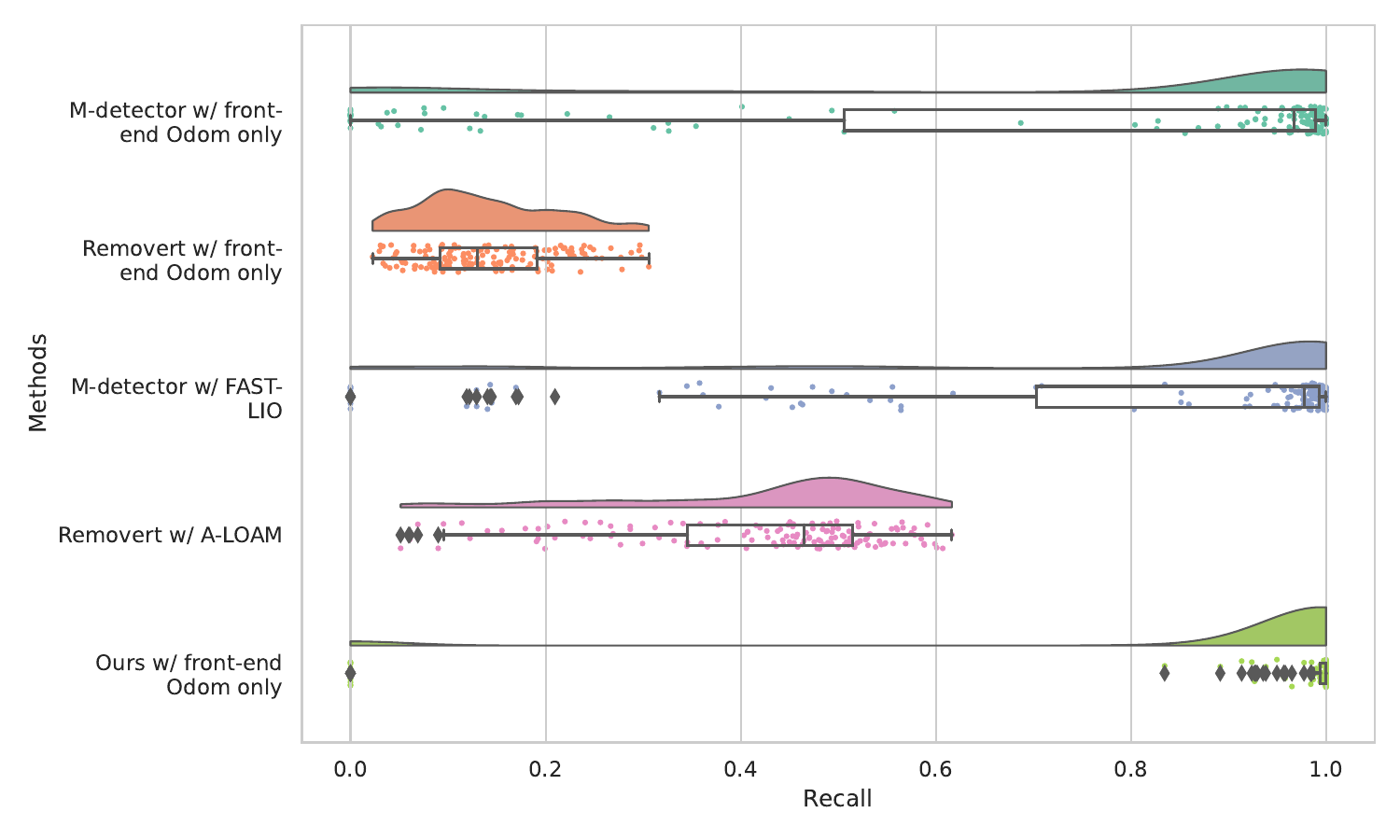}%
    \label{fig:recall2}%
  }

  \vspace{0.1cm} 

  \subfloat[\textit{Recall results in Sequence 3}: This subfigure presents the recall results for the third sequence. Our method consistently achieves the highest recall. M-detector with front-end odometry performs better than M-detector with FAST-LIO. Removert methods show similar lower recall trends.]{%
    \includegraphics[width=0.48\textwidth]{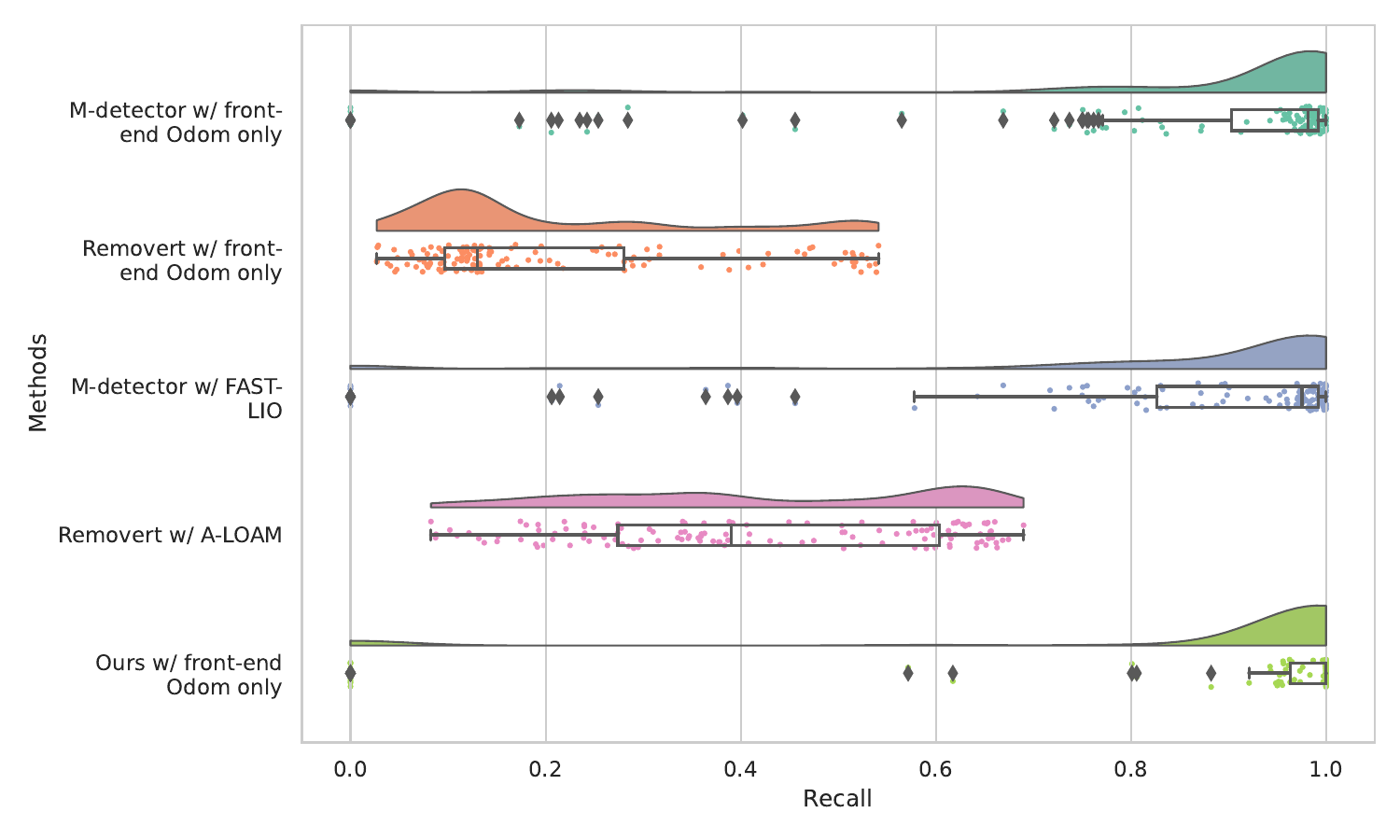}%
    \label{fig:recall3}%
  }\hfill
  \subfloat[\textit{Recall results in Sequence 4}: This subfigure illustrates the recall results for the fourth sequence. Our method leads with the highest recall and lowest variability. M-detector with FAST-LIO shows competitive recall but with occasional dips. M-detector with front-end odometry and Removert methods perform worse compared to previous sequences.]{%
    \includegraphics[width=0.48\textwidth]{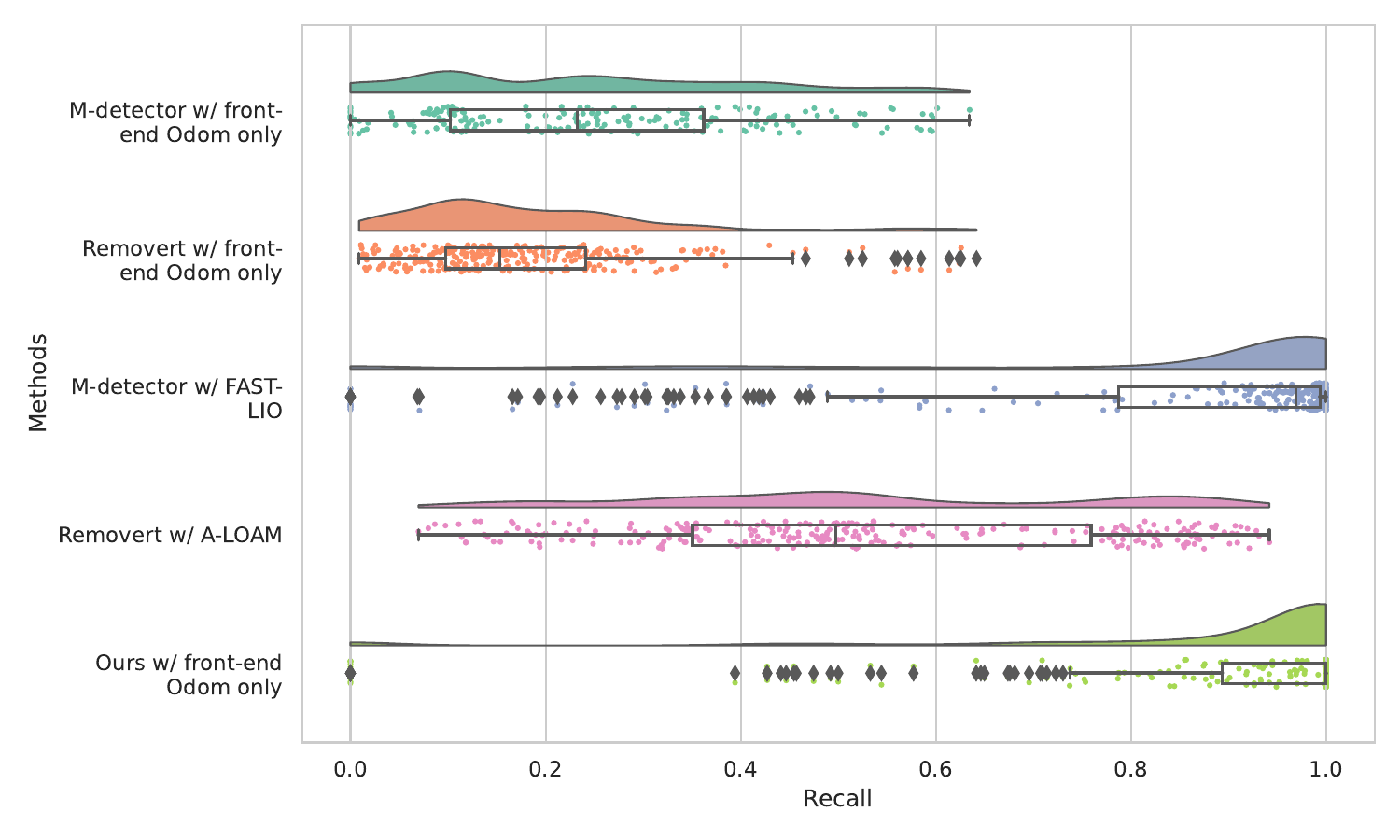}%
    \label{fig:recall4}%
  }
  \caption{\textbf{Recall results across four sequences}. This figure presents the recall results for different dynamic object detection methods across four test sequences. The methods compared are: M-detector with front-end odometry, Removert with front-end odometry, M-detector with FAST-LIO, Removert with A-LOAM, and our method with front-end odometry.}
  \label{fig:recall}
\end{figure*}

The recall results in Figure \ref{fig:recall} similarly highlight the robustness and effectiveness of our method. In Sequence 1, the recall performance of our method is notably similar to the M-detector with front-end odometry method. Among the methods, M-detector with FAST-LIO achieves the best recall results, demonstrating the highest recall values and the least variability. Despite this, our method remains highly competitive, exhibiting substantial recall performance that outperforms the other methods. Both our method and the M-detector with front-end odometry demonstrate superior recall results compared to the Removert methods, specifically Removert with front-end odometry and Removert with A-LOAM, which consistently show lower recall values and higher variability.

Sequence 2 recall results demonstrate that our method has the highest recall with minimal variability. M-detector with FAST-LIO shows high recall but notable fluctuations. M-detector with front-end odometry also exhibits competitive recall, although with more variability.
The lower recall and higher variability in Removert with A-LOAM and Removert with front-end odometry, indicate consistent poor performance across different sequences.

In Sequence 3, our method consistently achieves the highest recall, reinforcing its robustness. M-detector remains high in recall but less stable. However, in this sequence, the M-detector with front-end odometry has better performance than M-detector with FAST-LIO. Removert with A-LOAM and M-detector with front-end odometry show similar lower recall trends as previously observed.

The recall results in Sequence 4 further illustrate the effectiveness of our method, leading with the highest recall and lowest variability. M-detector with FAST-LIO shows competitive recall but with occasional dips. However, in this sequence M-detector with front-end odometry's recall performance is not as good as previous sequences. It has similir lower recall as Removert with front-end odometry . Removert with A-LOAM preformes better than before but still has lower recall compared to M-detector with FAST-LIO and our method.

\begin{figure*}[ht!]
  \centering
  \captionsetup[subfloat]{font=small, labelfont=bf}
  \subfloat[\textit{F1 Score results in Sequence 1}: This subfigure demonstrates our method with front-end odometry achieving the highest and most consistent F1 Score, significantly outperforming the other methods. M-detector with FAST-LIO shows competitive scores but with notable variability. Removert with A-LOAM and M-detector with front-end odometry exhibit lower F1 Scores and increased variability. Removert with A-LOAM shows slight better performance compared with other sequences.]{%
    \includegraphics[width=0.48\textwidth]{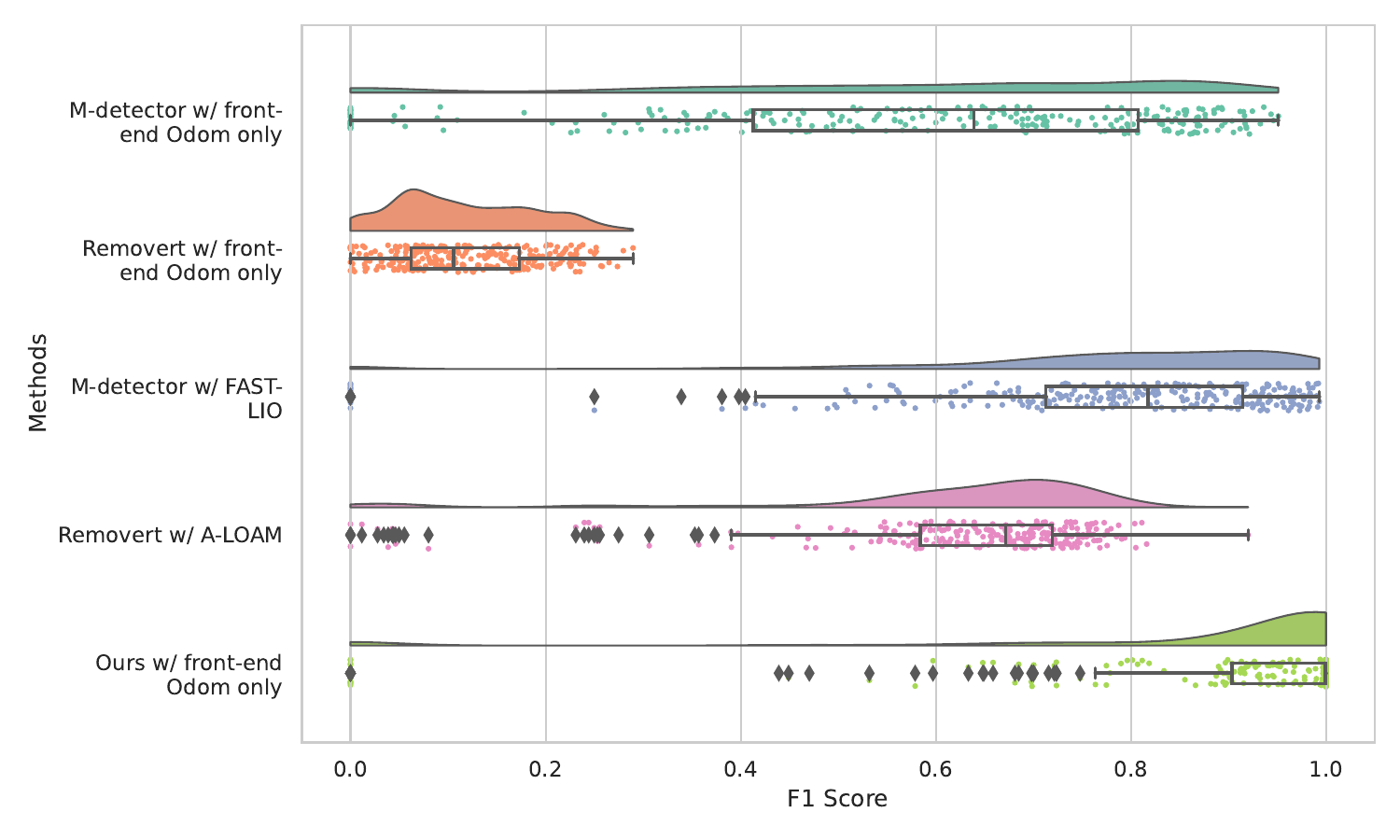}%
    \label{fig:f1_score1}%
  }\hfill
  \subfloat[\textit{F1 Score results in Sequence 2}: Our method continues to maintain the highest F1 Score with minimal variability, echoing the precision and recall outcomes. M-detector with FAST-LIO, although high, exhibits significant fluctuations. Both Removert and M-detector with front-end odometry confirm their weaker performance under noisy conditions by showing lower scores or higher variability.]{%
    \includegraphics[width=0.48\textwidth]{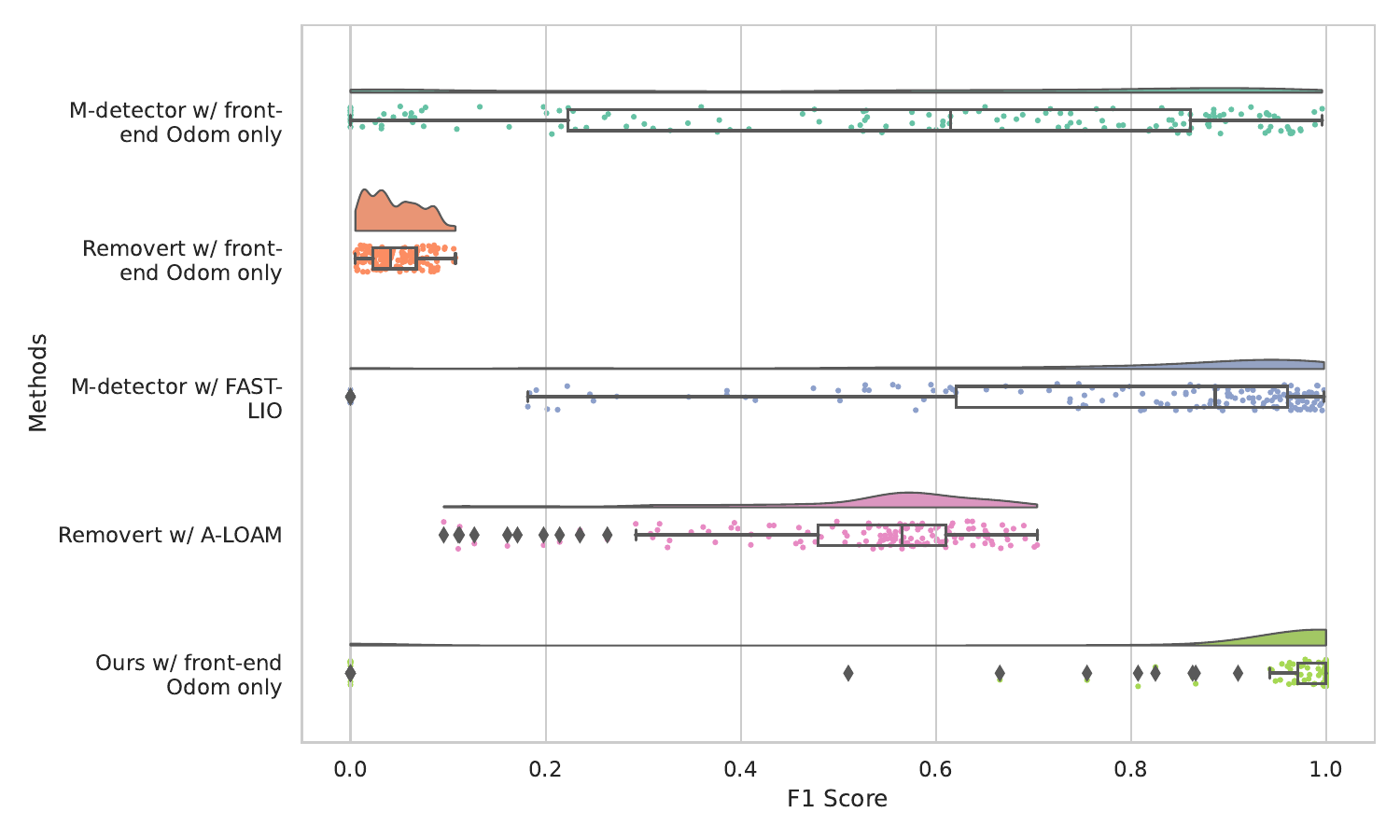}%
    \label{fig:f1_score2}%
  }
  \vspace{0.1cm} 
  \subfloat[\textit{F1 Score results in Sequence 3}: This subfigure highlights the robustness of our method, consistently achieving the highest F1 Score. M-detector with FAST-LIO remains competitive but but less stable. M-detector with front-end odometry outperforms Removert with A-LOAM in this sequence, though both continue to demonstrate high variability and lower scores.]{%
    \includegraphics[width=0.48\textwidth]{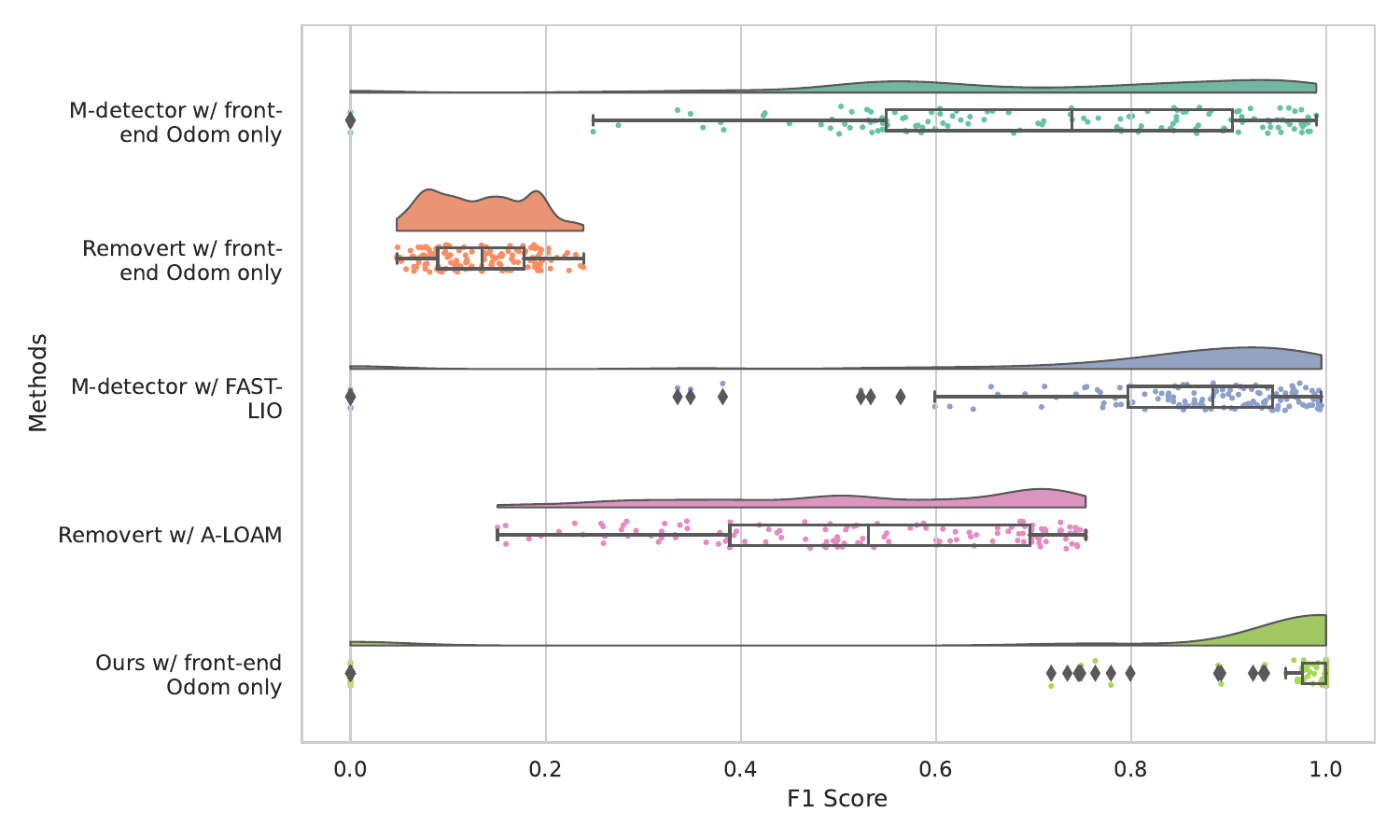}%
    \label{fig:f1_score3}%
  }\hfill
  \subfloat[\textit{F1 Score results in Sequence 4}: Our method leads with the highest F1 Score and lowest variability, confirming its consistent performance across sequences. M-detector with FAST-LIO remains competitive but with decreased stability. Removert with front-end odometry consistently shows the lowest F1 Scores, consistent with earlier patterns, while M-detector with front-end odometry displays similar scores but with increased variability.]{%
    \includegraphics[width=0.48\textwidth]{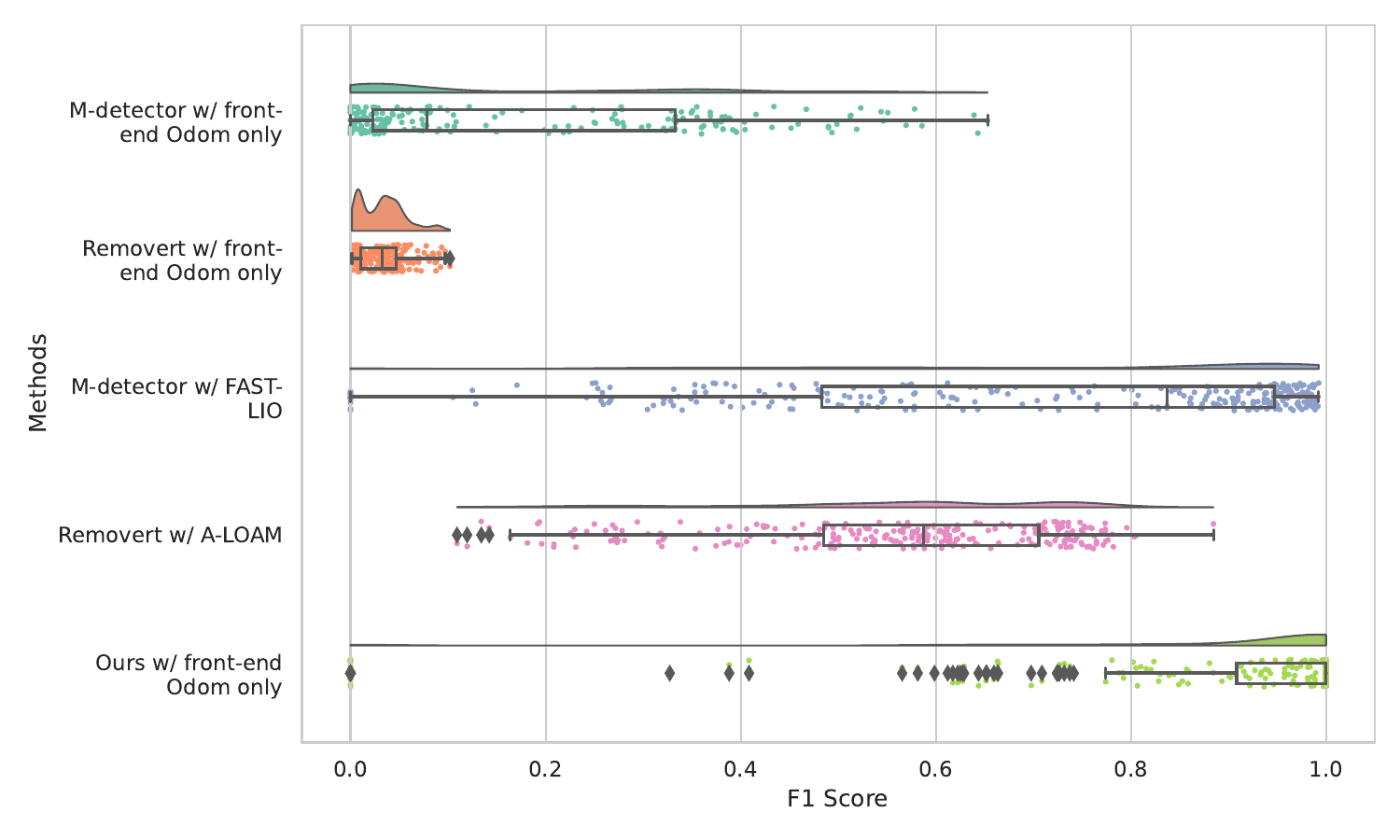}%
    \label{fig:f1_score4}%
  }
  \caption{F1 Score results across four sequences. This figure presents the F1 Score results for different dynamic object detection methods across four test sequences. The methods compared are: M-detector with front-end odometry, Removert with front-end odometry, M-detector with FAST-LIO, Removert with A-LOAM, and our method with front-end odometry.}
  \label{fig:f1_score}
\end{figure*}

The F1 Score results in Figure \ref{fig:f1_score} across the sequences provide a comprehensive view of our method's performance. In Sequence 1, our method with front-end odometry achieves the highest and most stable F1 Score, significantly outperforming other methods. M-detector with FAST-LIO shows good F1 Score but with more variability, while Removert with A-LOAM and M-detector with front-end odometry show lower F1 Score and M-detector with front-end odometry shows higher variability. Removert with A-LOAM shows better F1 score compared with the performance in other sequences. The Removert with front-end odometry has the lowest F1 Score among all methods.

For Sequence 2, our method maintains the highest F1 Score with minimal variability, similar to the precision and recall results. M-detector with FAST-LIO shows high F1 Score but notable fluctuations. The lower F1 Score in Removert with front-end odometry and the higher variability of M-detector with front-end odometry, further confirm their consistent poor performance during noisy odometry.

Sequence 3 F1 Score results reinforce our method's robustness, consistently achieving the highest F1 Score. M-detector with FAST-LIO remains competitive but less stable. The M-detector with front-end odometry has better F1 score than Removert with A-LOAM, but they all have higher variability. The lowest F1 Score in Removert with front-end odometry is consistent with previous observations.

In Sequence 4, our method once again leads with the highest F1 Score and lowest variability. M-detector with FAST-LIO shows competitive F1 Score but with even worse stability compared the performance in the previous sequences. Removert with front-end odometry continue to display the lowest F1 Score, aligning with the patterns seen in earlier sequences.  M-detector with front-end odometry has similar F1 Score as Removert with front-end odometry, but with higher variability.

Overall, the detailed analysis of experimental results across precision, IoU, recall, and F1 score consistently highlights the superiority and robustness of our method with front-end odometry. The consistent high performance and low variability of our method across different sequences underscore its reliability and effectiveness in dynamic object detection.

\mynew{Additionally, we performed statistical analysis using the Wilcoxon Signed-Rank Test \cite{wilcoxon1992individual} to evaluate the significance of the differences between the methods. We selected the Wilcoxon Signed-Rank Test because some of the data for precision, IoU, recall, and F1 score do not follow the normal distribution, and some data present outliers or extreme values, such as 0 or 1. Therefore, the Wilcoxon Signed-Rank Test is well-suited for this analysis. The results demonstrate that our method with front-end odometry significantly outperforms the other methods in terms of precision, IoU, recall, and F1 score, with all p-values $ < 0.05$. This statistical analysis further confirms the superior performance of our method in dynamic object detection.}

\newpage
\clearpage

\section{Biography Section}
 

\begin{IEEEbiography}[{\includegraphics[width=1in,height=1.25in,clip,keepaspectratio]{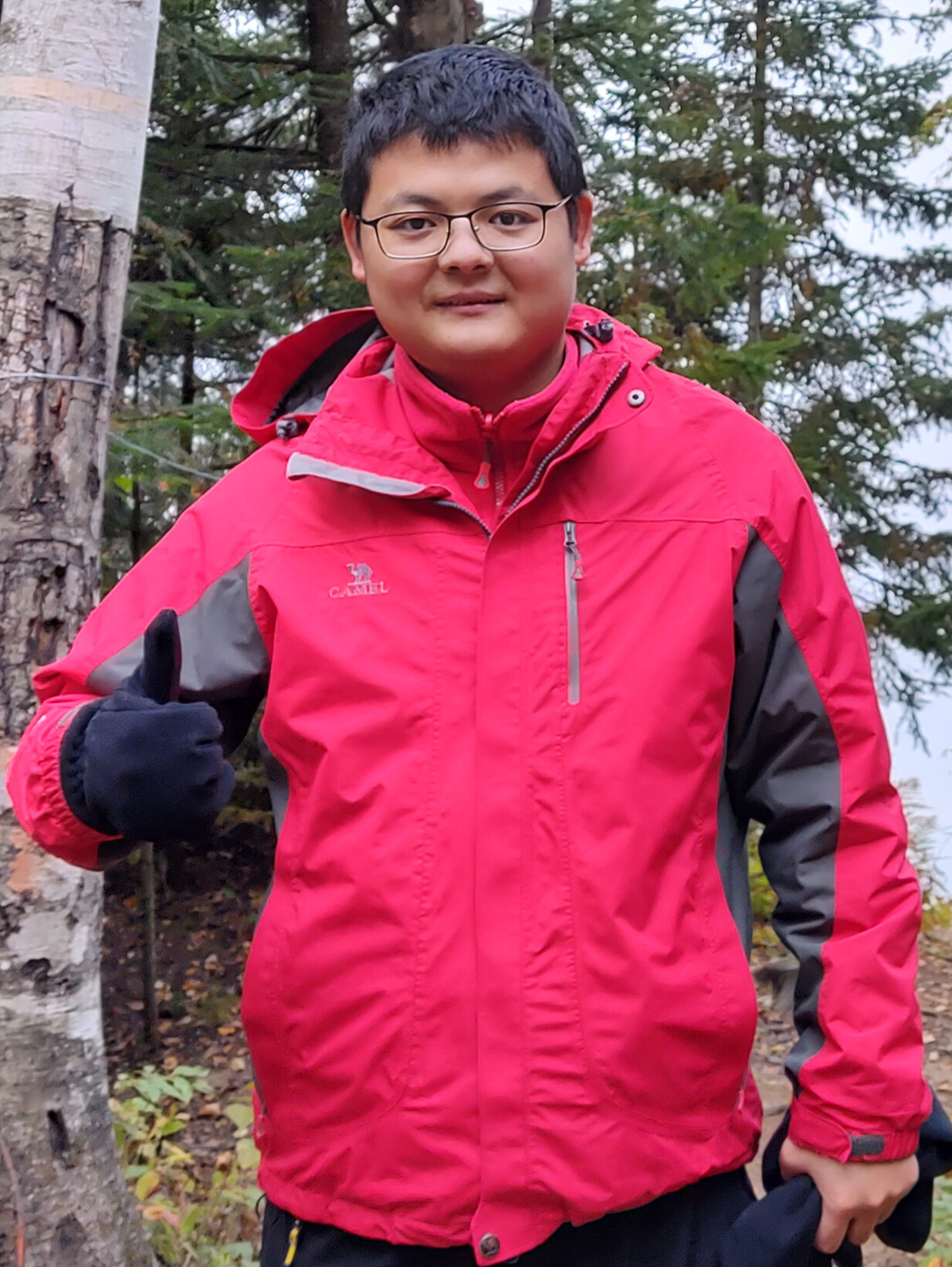}}]{Wenqiang Du} received the B.E. degree in the mechanical engineering from the Anhui University, Hefei, China in 2016, and the M.E. degree in the Pattern Recongnition and Intelligent System from the University of Chinese Academy of Sciences, Beijing, China in 2019. 

  He is currently working towards the Ph.D. degree in the MISTLab, the Department of Computer Engineering and Software Engineering, Polytechnique Montreal. His research interests include SLAM, robotics, perception, and dynamic objects detection.

\end{IEEEbiography}

\vspace{11pt}


\begin{IEEEbiography}[{\includegraphics[width=1in,height=1.25in,clip,keepaspectratio]{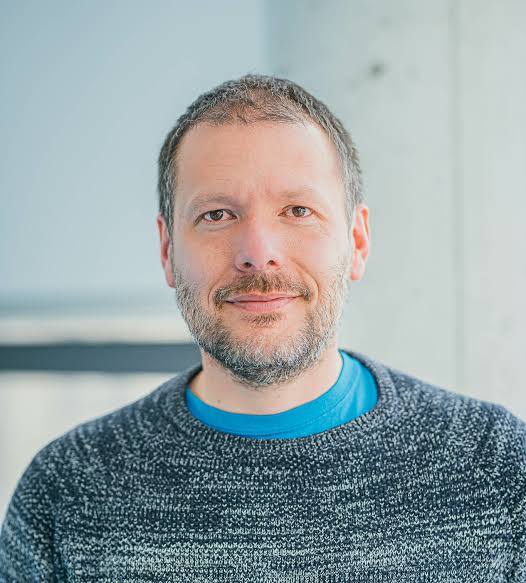}}]{Giovanni
    Beltrame}
  (Senior Member, IEEE) obtained his Ph. D. in Computer Engineering from
  Politecnico di Milano in 2006. He worked as microelectronics engineer at the
  European Space Agency on a number of projects, spanning from radiation
  tolerant systems to computer-aided design. Since 2010 he is Professor at
  Polytechnique Montreal with the Computer and Software Engineering Department,
  where he directs the MIST Lab and the ASTROLITH research group. He has
  authored or coauthored more than 100 papers in international journals and
  conferences. His research interests include modeling and design of embedded
  systems, artificial intelligence, and robotics.
\end{IEEEbiography}

\vfill

\end{document}